\documentclass{article}
\usepackage{PRIMEarxiv}
\usepackage{fancyhdr}
\usepackage[utf8]{inputenc}
\usepackage[T1]{fontenc}

\usepackage{amsmath}
\usepackage{amssymb}
\usepackage{amsfonts}

\usepackage{graphicx}
\graphicspath{{media/}}

\usepackage{booktabs}
\usepackage{multirow}
\usepackage{makecell}
\usepackage{diagbox}
\usepackage[table]{xcolor}
\usepackage{subcaption}
\usepackage{caption}
\usepackage{enumitem}
\usepackage{comment}
\usepackage{nicefrac}
\usepackage{microtype}

\usepackage{tikz}
\usetikzlibrary{matrix,positioning}

\usepackage{url}
\usepackage{orcidlink}
\usepackage{hyperref}

\title{Promptable Animal Pose Tracking Across Species
\thanks{\textit{Accepted for presentation at the ECCV 2026 Workshop on CV4Ecology.}}
}

\author{
  Le Li,
  Daniela Ivanova,
  Nicolas Pugeault\\[0.8em]
  School of Computing Science \\[0.2em]
  University of Glasgow \\[0.6em]
  \texttt{3120568L@student.gla.ac.uk}\\
  \texttt{\{Daniela.Ivanova, Nicolas.Pugeault\}@glasgow.ac.uk}
}

\begin{document}
\maketitle
\thispagestyle{empty}

\begin{abstract}
Animal pose estimation and tracking is important for wildlife monitoring and conservation research, and with limited expert time for labelling automated approaches are imperative. While human pose estimation and tracking has seen rapid progress thanks to large annotated datasets, animal pose remain challenging, due to large morphological and behavioural differences between species and limited annotated data. Existing approaches either optimise generic keypoint localisation from annotated datasets (such as APTv2) with poor generalisation, or track custom keypoints using visual tracking, at the cost of performance. In this paper, we demonstrate that vision foundation models trained on large datasets can be used effectively to track animal pose with limited labelled data. We propose two models, one unsupervised and the other supervised, to track user-selected keypoints in videos. The supervised approach delivers superior tracking accuracy by employing a keypoint prompt encoder to explicitly inject structural priors from a reference frame into feature matching. In parallel, the unsupervised route provides strong cross-species robustness by leveraging diverse foundation-model features for training-free correspondence matching. Extensive evaluation on challenging animal video benchmarks APTv2 and TigDog demonstrates that our framework achieves strong performance while maintaining an effective balance between accuracy and generalisation, offering a practical solution for real-world animal behaviour analysis and conservation applications.
\end{abstract}


\section{Introduction}
\label{sec:intro}
Animal pose estimation and tracking play an important role in livestock monitoring~\cite{perneel2025consistent, li2023real}, wildlife research~\cite{han2024multi, mcnutt2024whole}, and conservation~\cite{zhang2020omni, deng2024towards}. Animal Pose Estimation (APE) aims to localize keypoints from individual images, whereas Animal Pose Tracking (APT) aims to localize and track animal keypoints across video sequences.  APT serves as a fundamental step for downstream tasks such as animal action recognition, and is therefore crucial for analysing animal behaviour in both laboratory and in-the-wild settings.

\begin{figure}[!t]
	\centering
	\includegraphics[width=\linewidth]{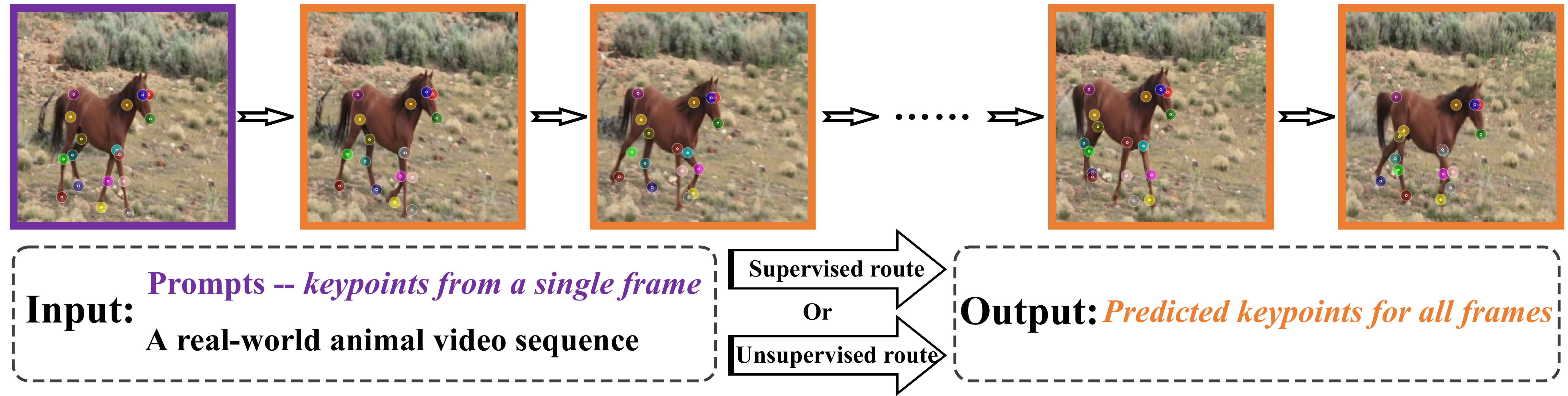}
	\caption{\centering A concise overview of our proposed promptable animal pose tracking pipeline. The supervised route focuses on accurate correspondence learning, while the unsupervised route emphasizes cross-species robustness.}
	\label{fig1}
\end{figure}

Despite recent progress in APT, existing approaches are limited in their applicability and often rely heavily on expensive manual supervision. In terms of application scope, methods for tracking animal keypoints across video frame sequences primarily focus on laboratory animals~\cite{lauer2021multi, lauer2022multi, pereira2020sleap, pereira2019fast} or wild animals in specialized 3D environments~\cite{waldmann2022muppet, waldmann20243d}. Apart from the constraint of being primarily designed and evaluated in controlled environments, some of these approaches additionally require modalities beyond standard RGB input~\cite{waldmann20243d}.  

Meanwhile, from a methodological perspective, existing approaches often rely heavily on large-scale densely annotated supervision~\cite{kresovic2022pigpose, choi2026adoption}. While some studies have explored annotation-efficient alternatives, such as synthetic data~\cite{mu2020learning, li2021synthetic, shooter2024sydog}, few-shot, and zero-shot paradigms~\cite{li2023scarcenet, ye2024superanimal}, these approaches typically suffer from suboptimal accuracy. For APT, behaviour analysis across multiple species and wildlife conservation demand a flexible solution that can generalize to diverse real-world scenarios from simple RGB videos while reducing the reliance on expensive annotations without sacrificing tracking accuracy.

Beyond limitations in application scope and annotation efficiency, existing approaches still fall short in meeting the practical needs of animal researchers. In real-world settings, researchers often require flexible annotation and tracking tools rather than large-scale supervised models with fixed predefined keypoints. Ideally, such tools should allow users to specify only a few keypoints of interest, support stronger cross-species generalization, and reduce dependence on large-scale annotations. These requirements motivate a shift from conventional human pose estimation frameworks toward foundation-model-based approaches for keypoint correspondence and tracking.

In recent years, foundation models have achieved remarkable progress in visual representation learning. Vision transformers such as DINOv3~\cite{simeoni2025dinov3} have demonstrated strong semantic correspondence and cross-domain generalization capabilities, while diffusion-model-based representations~\cite{luo2023diffusion, stracke2025cleandift} have shown impressive performance in dense matching and visual correspondence tasks. Their effectiveness has also been demonstrated for point tracking through video-specific optimization in DINO-Tracker~\cite{tumanyan2024dino}. Despite these advances, the potential of directly leveraging general foundation-model representations for animal pose tracking remains largely unexplored.

To address the aforementioned issues, we propose a flexible promptable pipeline for animal keypoint tracking in arbitrary RGB video sequences. As illustrated in Figure \ref{fig1}, our method takes a video sequence together with keypoint annotations from a single reference frame as input. Building upon diverse foundation-model features, we introduce a novel promptable animal pose tracking framework that can accurately predict keypoints across all remaining frames through either a supervised or an unsupervised route. Levaraging foundation-model features extracted from video frames, the supervised route incorporates a novel keypoint encoder to integrate annotated keypoint priors into the reference-frame representations. The enhanced features are subsequently processed by a matcher module, where the keypoint encoder and matcher are jointly trained for accurate supervised animal pose tracking. In contrast, the unsupervised route directly establishes dense pixel-level correspondences between foundation-model features under bounding-box constraints, enabling training-free localization of corresponding keypoints throughout the video sequence. As shown in Table \ref{taba1} of the Appendix, our approach follows a fundamentally different paradigm from conventional frameworks, enabling a balance between tracking accuracy and annotation efficiency across the two routes.

The contributions of this paper are as follows:

\begin{itemize}
    \item A novel Promptable Animal Pose Tracking framework that leverages diverse pretrained foundation-level visual representations, rather than relying on conventional annotation-intensive pose estimation pipelines. 
    \item A supervised Animal Pose Tracking model that integrates a keypoint encoder with pretrained foundation-model representations, enabling the model to better focus on the structural priors provided by the input reference keypoints. 
    \item An unsupervised Animal Pose Tracking model that combines pretrained visual feature representations with a lightweight matcher equipped with drift-correction constraints, enabling robust tracking of annotated keypoints without any task-specific training. 
\end{itemize}

\section{Related Works}

The animal pose analysis field has evolved from Animal Pose Estimation (APE) in individual images to Animal Pose Tracking (APT) in video sequences. We trace representative approaches of APE (Section \ref{sec2_1}) and APT (Section \ref{sec2_2}), including current state-of-the-art. 

\subsection{Animal Pose Estimation}
\label{sec2_1}
Animal Pose Estimation (APE) poses unique challenges beyond those encountered in human pose estimation, including: diverse body topologies, large cross-species appearance variations, and expensive keypoint annotation. To tackle these issues, existing APE methods focus on bridging two major domain gaps: the \emph{human–animal} gap and the \emph{synthetic-to-real} gap. To address the lack of labeled data, several approaches have explored the use of synthetic data~\cite{mu2020learning, li2021synthetic, shooter2024sydog, bonetto2026zebrapose, li2023scarcenet}. With regards to backbone design, common APE methods largely follow established human pose estimation pipelines which build upon foundational architectures and frameworks, e.g. ResNet~\cite{he2016deep}, Stacked Hourglass~\cite{newell2016stacked}, OpenPose~\cite{cao2019openpose}, and HRNet~\cite{wang2020deep}. These models are directly transferred or fine-tuned to animal domains~\cite{pereira2019fast, ye2024superanimal, yu2021ap, cao2019cross}.While effective for single-image pose estimation, these designs are less suitable for video-based keypoint correspondence, which requires more flexible feature representations.

Recently, vision foundation models, such as DINOv3~\cite{simeoni2025dinov3}, diffusion-based representations like Diffusion Hyperfeatures~\cite{luo2023diffusion}, and multimodal models such as BioCLIP~\cite{stevens2024bioclip}, have demonstrated strong generalizable visual representations, and have benefited many downstream tasks such as human pose estimation~\cite{feng2023diffpose, chen2025diffusion} and action recognition~\cite{guimaraes2026diffusion, lu2025dposer}. Motivated by these advances, we hypothesise that APE can also benefit from large-scale vision foundation models. In this work, our model adopts general vision foundation models to extract dense per-pixel features for accurate keypoint matching, demonstrating the effectiveness of such representations in animal keypoint estimation and tracking.

\subsection{Animal Pose Tracking}
\label{sec2_2}

Animal Pose Tracking (APT) aims to track keypoints over video sequences for modeling temporal pose dynamics. However, apart from a few methods using temporal encoding~\cite{russello2022t}, most existing approaches still rely on frame-wise APE~\cite{mathis2018deeplabcut, lauer2022multi, pereira2020sleap, waldmann2022muppet, waldmann20243d}. Among them, DeepLabCut~\cite{mathis2018deeplabcut} introduces a method for labelling animal body parts and training a deep neural
network for predicting 2D body part positions. A follow-up work~\cite{lauer2022multi} supports multi-animal tracking across species such as mice, marmosets, and fish, combining SORT~\cite{bewley2016simple}-based local tracking with a global optimization step to resolve occlusions and interactions. Meanwhile, Pan et al.~\cite{pan2025animal} proposes a test-time optimization framework that adapts a general-purpose point tracker to track animals. Concurrently, general point tracking methods such as CoTracker~\cite{karaev2025cotracker3}, TAPIR~\cite{doersch2023tapir}, AllTracker~\cite{harley2025alltracker} and DINO-Tracker~\cite{tumanyan2024dino} have demonstrated strong capability in tracking arbitrary points across long video sequences, and can be applied to animal videos. These methods rely on large-scale tracking supervision and perform video-specific optimization, representing generalised solutions for point tracking.

Although effective, existing approaches dedicated to APT are restricted primarily to fully supervised settings~\cite{yang2022apt, mathis2018deeplabcut}. Most rely on fixed skeletal priors with poor cross-species generalization. Due to the nature of real-world applications, APT often requires flexible attention mechanisms and the ability to operate with limited or even no annotations. To address this limitation, our method provides both supervised and unsupervised pipelines, enabling keypoint tracking across video sequences from a single annotated frame, thereby reducing annotation cost while maintaining competitive performance.

\section{Method}
\label{sec3}

Our promptable animal pose tracking pipeline leverages foundation-model features for robust cross-frame correspondence, and is implemented via both supervised and unsupervised routes. In both settings, the model takes an animal video sequence, together with keypoint annotations from a single reference frame as input. The output of the model is keypoint trajectories across all frames. A single RGB animal video can be represented as $\mathcal{V}=\{{I}^{t}\, |\, t=1, \ldots, T\}$, while ${I}^{t}$ is the frame image at time ${t}$. The frame selected for input keypoint annotation is referred to as the reference frame, which is denoted as ${I}^{r}$. And the corresponding keypoint annotations are represented as $\mathcal{P}^r = \{p_k^r \mid k = 1, \ldots, K\}$, where each keypoint location is defined as $p_k^r = (x_k^r, y_k^r)$. Here, $K$ denotes the total number of keypoints, and (${x}_{k}^{r}$, ${y}_{k}^{r}$) represents the 2D spatial coordinates of the $k$-th keypoint in the reference frame. The objective is to estimate the keypoint locations across all frames given the reference frame, formulated as $\hat{\mathcal{P}} = \{\hat{p}_k^{t} \mid k = 1, \ldots, K,\; t = 1, \ldots, T,\; t \neq r\}$. To achieve this, we first pass the input frame sequence $\mathcal{V}$ through a foundation model to extract per-frame features (Section \ref{sec3_1}). Then, to emphasize either accuracy or cross-class generalization, the method branches into an unsupervised route (Section \ref{sec3_2}) and a supervised route (Section \ref{sec3_3}).

\subsection{Frame Feature Extraction}
\label{sec3_1}

Given a video $\mathcal{V}$, our model takes a reference frame ${I}^{r}$, keypoint annotations $\mathcal{P}^{r}$, and a target frame ${I}^{t}$, where ${I}^{t}$ is sampled from all frames in $\mathcal{V} \setminus \{ I^{r} \}$. In both the unsupervised and supervised routes, feature extraction for ${I}^{r}$ and ${I}^{t}$ is performed as the first step:

\begin{equation}
\mathcal{F}^{r}_{pre} = \psi(I^{r}), \quad
\mathcal{F}^{t}_{pre} = \psi(I^{t})
\end{equation}

where $\mathcal\psi$ denotes the feature extractor, and $\mathcal{F}^{r}_{pre} \in \mathbb{R}^{H \times W \times C}$ and $\mathcal{F}^{t}_{pre} \in \mathbb{R}^{H \times W \times C}$ denote dense per-pixel feature maps. Since we adopt general vision foundation models, we have multiple choices for representing the two frames such as DINOv3~\cite{simeoni2025dinov3}, BioClip~\cite{stevens2024bioclip}, CleanDIFT~\cite{stracke2025cleandift}, and Diffusion Hyperfeatures~\cite{luo2023diffusion}. Through extensive experiments, we find that Diffusion Hyperfeatures provide the most effective representation for dense correspondence in our setting. We therefore adopt them as our primary feature extractor for cross-frame keypoint correspondence.

\subsection{Unsupervised Route}
\label{sec3_2}

\begin{figure}[!h]
	\centering
	\includegraphics[scale=0.7]
    {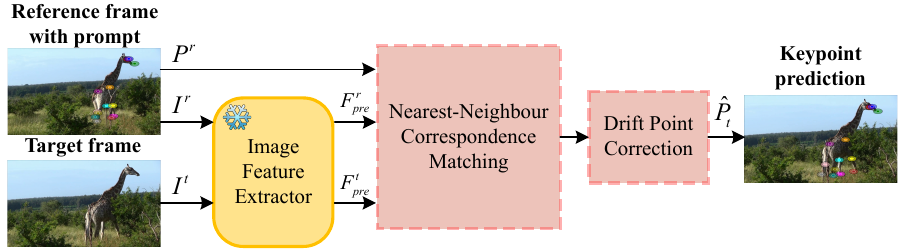}
	\caption{\centering Overall pipeline of the unsupervised route.}
	\label{fig3}
\end{figure}

As depicted in Figure \ref{fig3}, the unsupervised route employs a corresponding matching of $\mathcal{F}^{r}_{pre}$ and $\mathcal{F}^{t}_{pre}$ with a novel drift point correction module, and is entirely training-free. The pipeline consists of two steps.

\noindent\textbf{Step 1. Nearest-Neighbour matching}

After extracting the corresponding features $\mathcal{F}^{r}_{pre}$ and $\mathcal{F}^{r}_{pre}$ from ${I}^{r}$ and ${I}^{t}$, the method directly performs Nearest-Neighbour matching to localize the keypoints in the target frame:

\begin{equation}
\hat{\mathcal{P}}^{t}
=
\mathrm{NN}\big(\mathcal{F}^{r}, \mathcal{F}^{t}, \mathcal{P}^{r}\big),
\quad
\hat{\mathcal{P}}^{t}
=
\{\hat{p}_k^{t}\}_{k=1}^{K}
\end{equation}

Specifically, each target keypoint of each video frame is determined by:

\begin{equation}
\hat{p}_k^{t}
=
\arg\max_{p \in \Omega}
\left\langle
\mathcal{F}^{r}(p_k^{r}),
\mathcal{F}^{t}(p)
\right\rangle
\end{equation}

\noindent\textbf{Step 2. Drift points correction with bounding box constraint.}

In the basic setting, $\Omega$ denotes the set of all candidate spatial locations in the target frame, corresponding to all pixel locations across the entire image.

However, due to the characteristic body structures of many quadrupedal mammals, such as elongated limbs with highly similar appearances across individuals, keypoints can easily drift from one animal to another when multiple animals appear together. To mitigate this issue, we introduce a bounding-box constraint that restricts correspondence predictions within the appropriate object region:

\begin{equation}
\hat{\mathcal{P}}^{t}
=
\{\tilde{p}_k^{t}\}_{k=1}^{K},
\quad
\tilde{p}_k^{t}
=
\begin{cases}
\hat{p}_k^{t}, & \hat{p}_k^{t} \in \Omega_k^{\mathrm{bbox}} \\
\arg\max\limits_{p \in \Omega_k^{\mathrm{bbox}}}
\left\langle
\mathcal{F}^{r}(p_k^{r}),
\mathcal{F}^{t}(p)
\right\rangle,
& \text{otherwise}
\end{cases}
\end{equation}

The box-constrained candidate region $\Omega_k^{\mathrm{bbox}}$ is defined based on the bounding boxes provided by the dataset and the process of drift points correction is entirely optional. We denote the sets of instance bounding boxes in the reference and target frames as $\mathcal{B}^{r} = \{B_m^{r}\}_{m=1}^{M_r}$ and $\mathcal{B}^{t} = \{B_n^{t}\}_{n=1}^{M_t}$, where each box is represented as $B=(x_{\min}, y_{\min}, x_{\max}, y_{\max})$. However, since instance identities are inherently ambiguous and may change across frames in multi-animal video sequences, we do not assume explicit box-level tracking between frames. Instead, we establish a soft correspondence by matching instances based on the spatial proximity of their box centres. 

Specifically, for each reference box $B_m^{r}$, we assign a target box $B_{n_m^{*}}^{t}$ via distance-based matching in the centre space:

\begin{equation}
n_m^{*} = \arg\min_{n} \|c_m^{r} - c_n^{t}\|_2,
\end{equation}

where $n_m^{*}$ denotes the index of the selected target box for the $m$-th reference box. $c_m^{r}$ and $c_n^{t}$ represent the geometric centres of the corresponding boxes, and $m,n$ index reference and target instances respectively.

To associate keypoints with instance-level structure, we identify all reference boxes that contain a given keypoint $p_k^{r}$, forming a set $\mathcal{M}_k$.

For a keypoint $p_k^{r}$, the valid search region in the target frame is constructed by transferring the corresponding reference instances to the target frame via the above box matching. The region is defined as the union of the matched target boxes:

\begin{equation}
\Omega_k^{\mathrm{bbox}} = \bigcup_{m \in \mathcal{M}_k} B_{n_m^{*}}^{t}.
\end{equation}

Thus, $\Omega_k^{\mathrm{bbox}}$ represents a spatially constrained candidate region in the target frame for the $k$-th keypoint, restricting correspondence search to instance-consistent areas.

\subsection{Supervised Route}
\label{sec3_3}

\begin{figure}[!h]
	\centering
	\includegraphics[width=\linewidth]{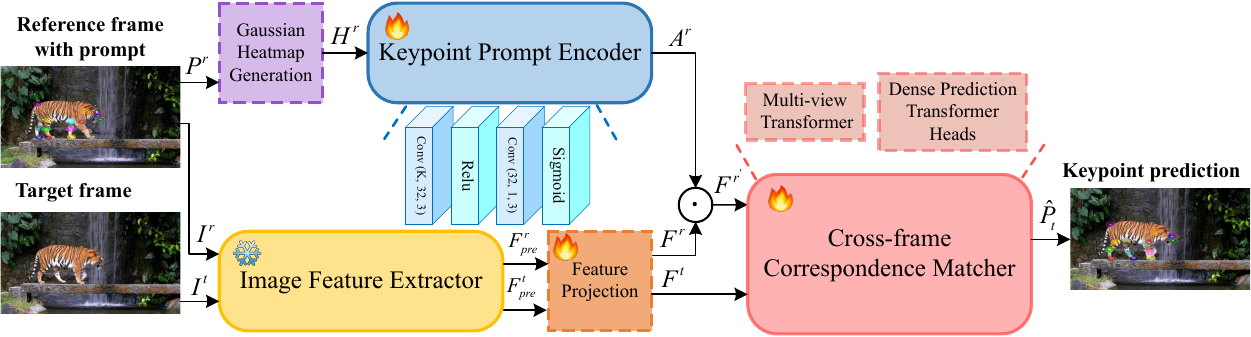}
	\caption{\centering Overall framework of the supervised route. The keypoint prompt encoder, feature projection, and cross-frame correspondence matcher are fine-tuned. $\odot$ denotes element-wise multiplication.}
	\label{fig2}
\end{figure}

The supervised pipeline is depicted in Figure \ref{fig2}. It builds upon the simplified unsupervised version, and is designed to handle tasks requiring higher precision. The entire pipeline could be divided into 5 steps.

\noindent\textbf{Step 1. Feature extraction and feature projection}

The same as in the unsupervised route, we obtain pre-extracted features for each frame $\mathcal{F}^{r}_{pre}$ and $\mathcal{F}^{t}_{pre}$. A convolutional projection module is then used to adapt the extracted features to the target task and enhance their representational capacity, producing compact feature representations for subsequent matching:

\begin{equation}
\mathcal{F}^r = \phi(\mathcal{F}^{r}_{pre}), \quad
\mathcal{F}^t = \phi(\mathcal{F}^{t}_{pre})
\end{equation}

where $\phi$ is the learnable convolutional projection.

\noindent\textbf{Step 2. Keypoint heatmap construction and feature encoding}

The input keypoint annotations $\mathcal{P}^{r}$ of the reference frame ${I}^{r}$ are firstly converted into dense Gaussian heatmaps to provide a smoother prompt for the model:

\begin{equation}
\mathcal{H}^r = g(\mathcal{P}^r)
\end{equation}

Specifically, the keypoint heatmaps are denoted as $\mathcal{H}^r = \{\mathcal{H}_k^r \mid k = 1, \ldots, K\}$, where each heatmap $\mathcal{H}_k^r \in \mathbb{R}^{H \times W}$ is generated from the corresponding keypoint location $p_k^r$ as:
\begin{equation}
		\label{eq1}
		\mathcal{H}_k^r(u,v) = \exp\left(-\frac{(u-y_k^r)^2 + (v-x_k^r)^2}{2\sigma^2}\right)
\end{equation}
where $(u,v)$ denotes a pixel location on the heatmap, and $\sigma$ controls the Gaussian spread around each annotated keypoint. Thus, the complete keypoint prompt can be represented as $\mathcal{H}^r \in \mathbb{R}^{K \times H \times W}$ where $\mathcal K$ occupies a separate channel.

After converting the input keypoint annotations $\mathcal{P}^{r}$ into Gaussian heatmaps $\mathcal{H}^{r}$, they are passed through a lightweight Keypoint Prompt Encoder to produce a single structural attention map:

\begin{equation}
\mathcal{A}^r = E_{kp}(\mathcal{H}^r)
\end{equation}

where $\mathcal{A}^{r}$ is the resulting attention map. And it can be denoted as $\mathcal{A}^r \in \mathbb{R}^{1 \times H \times W}$.

\noindent\textbf{Step 3. Keypoint-guided feature modulation}

Finally, the attention map $\mathcal{A}^{r}$ is used to modulate the feature representation of reference frame $\mathcal{F}^{r'}$, in order to enhance regions corresponding to the $K$ keypoints and encouraging the model to focus more explicitly on keypoint-relevant spatial structures: 

\begin{equation}
\mathcal{F}^{r'} = \mathcal{F}^{r} \odot \left(1 + \alpha \cdot A^{r}\right)
\end{equation}

where $\odot$ stands for element-wise multiplication, and $\alpha$ represents a learnable scalar.

\noindent\textbf{Step 4. Cross-frame correspondence estimation}

We adopt a coarse feature matching strategy inspired by RoMav2~\cite{edstedt2025roma} to establish dense cross-frame correspondences:
\begin{equation}
\hat{\mathcal{P}}_{t}, \; \mathcal{C}_{t}
= M_{\theta}\big(\mathcal{F}^{r'}, \mathcal{F}^{t}\big)
\end{equation}

where $M_{\theta}$ denotes the matcher for cross-frame correspondence estimation. $\hat{\mathcal{P}}_{t}$ and $\mathcal{C}_{t}$ represent the keypoint prediction and confidence of ${I}^{t}$.

\noindent\textbf{Step 5. Optimization objective and loss function}

In conclusion, the objective of the supervised route is to predict the target keypoint locations:

\begin{equation}
\hat{\mathcal{P}} = \{ f_{\theta}(I^{r}, \mathcal{H}^{r}, I^{t}) \mid I^{t} \in \mathcal{V} \setminus \{ I^{r} \} \}
\end{equation}

where $f_{\theta}$ denotes the proposed tracking model parameterized by $\theta$, and $\hat{\mathcal{P}}$ represents the predicted keypoints for video ${V}$.

And the loss can be defined as:

\begin{equation}
\mathcal{L}
= \frac{1}{N}
\sum_{i=1}^{N}
\sum_{k=1}^{K}
m_{ik} \left\|
\hat{p}_{ik}^t - p_{ik}^t
\right\|_2^2,
\quad
m_{ik} \in \{0,1\}
\end{equation}

where the mask $m_{ik}$ marks visibility of a keypoint.
\section{Experiments}
\label{sec4}

We conduct detailed comparative and ablation studies under both unsupervised and supervised settings on two large-scale Animal Pose Tracking (APT) datasets: APTv2~\cite{yang2023aptv2} and TigDog~\cite{del2017behavior}. Overall, the results demonstrate that our proposed Promptable APT method achieves strong tracking performance under supervised route and generalization ability under unsupervised route.

\subsection{Experimental Setup}
\label{sec4_1}
\subsubsection{Datasets}
We evaluate on two large-scale animal video datasets, APTv2~\cite{yang2023aptv2} and TigDog~\cite{del2017behavior}. Both datasets consist of temporally organized video sequences, and are collected from diverse real-world scenarios, featuring natural pose variations and visually complex backgrounds with realistic noise.

\textbf{APTv2} is the largest annotated animal video dataset for keypoint-based pose tracking. It contains 30 animal species across 15 families; each video is of fixed length of 15 frames. In our setting, keypoints annotated in a single reference frame are propagated to the remaining 14 frames for evaluation. All experiments reported in this paper use the first frame as the reference frame for consistency. However, the proposed method is not restricted to this choice.

\textbf{TigDog} is a widely used video dataset for APT. Although the species coverage is limited to only tigers and horses, TigDog remains challenging due to its diverse real-world backgrounds. It comprises longer sequences without a fixed length, which introduces additional complexity to keypoint tracking over extended temporal ranges.

\subsubsection{Implementation Details}
All training and testing are conducted using a device equipped with a Quadro RTX 6000 GPU. We evaluate DINOv3~\cite{simeoni2025dinov3}, BioCLIP~\cite{stevens2024bioclip}, CleanDIFT~\cite{stracke2025cleandift}, and Diffusion Hyperfeatures~\cite{luo2023diffusion} as feature extractors. For the supervised route, we compare DINOv3 and Diffusion Hyperfeatures. Our matcher is adapted from the coarse matching module of RoMa v2~\cite{edstedt2025roma}. Additional implementation details are provided in Appendix B.

We follow the standard APT evaluation metric, Percentage of Correct Keypoints (PCK), which measures the proportion of predicted keypoints within a predefined normalized distance from the ground truth. We report image-normalized PCK (PCK@0.1$_{\text{img}}$ and PCK@0.05$_{\text{img}}$), where distances are normalized by the image resolution.

\subsection{Results on the unsupervised route}
\label{sec4_2}

\subsubsection{Overall Analysis}

Table \ref{tab_unsuper_all} presents the performance of our unsupervised route on APTv2 and TigDog under different backbones and settings. Overall, diffusion-based features achieve the strongest performance. Since diffusion models support textual conditioning during denoising, we further evaluate species-name prompts at inference time. “w prompt on species type” uses the species name as the prompt, whereas “w/o prompt on species type” uses an empty prompt. Both diffusion-based methods consistently benefit from species-name prompts under both evaluation metrics.

\begin{table*}[ht!]
\centering
\resizebox{\textwidth}{!}{
\begin{tabular}{llcccc}
\toprule
\multirow{2}{*}{\textbf{Method}} & 
\multirow{2}{*}{\textbf{Implementation}} & 
\multicolumn{2}{c}{\textbf{APTv2}} & 
\multicolumn{2}{c}{\textbf{TigDog}} \\
\cmidrule(lr){3-4}\cmidrule(lr){5-6}
 & & \textbf{PCK@0.1$_{\text{img}}$} & \textbf{PCK@0.05$_{\text{img}}$} & \textbf{PCK@0.1$_{\text{img}}$} & \textbf{PCK@0.05$_{\text{img}}$} \\
\hline 

Ours (BioClip) 
 & Last-4-layer concatenation & 42.9 & 21.9 & 22.7 & 11.8 \\
\addlinespace

Ours (DINOv3) 
 & Last-layer patch features & 67.9 & 42.4 & 50.8 & 29.3 \\
 & All-layer concatenation & 69.4 ($\uparrow$1.5) & 43.9 ($\uparrow$1.5) & 51.3 ($\uparrow$0.5) & 29.6 ($\uparrow$0.3) \\
\addlinespace

Ours (CD) 
 & w/o prompt on species type & 82.6 & 60.6 & 72.4 & 51.4 \\
 & w prompt on species type & 84.0 ($\uparrow$1.4) & 61.9 ($\uparrow$1.3) & 75.1 ($\uparrow$2.7) & 53.5 ($\uparrow$2.1) \\
\addlinespace

\begin{tabular}[t]{@{}l@{}}Ours (DHf)\end{tabular}
 & w/o prompt on species type & 85.1 & 68.6 & 81.3 & 67.4 \\
 & w prompt on species type & \underline{86.6} ($\uparrow$1.5) & \underline{70.5} ($\uparrow$1.9) & \underline{83.0} ($\uparrow$1.7) & \underline{69.7} ($\uparrow$2.3) \\
\bottomrule
\end{tabular}}
\caption{Unsupervised results on APT-v2 and TigDog with different backbones. The results on TigDog do not use the Drift Points Correction module. $\uparrow$ denotes performance improvement relative to the previous row. CD denotes CleanDIFT, and DHf denotes Diffusion Hyperfeatures. The best result is \underline{underlined}.}
\label{tab_unsuper_all}
\end{table*}

\begin{figure*}[ht]
\centering

\begin{minipage}[t]{0.4\textwidth}
\vspace{0pt}
\centering
\resizebox{\linewidth}{!}{
\begin{tabular}{lcc}
\toprule
\multirow{2}{*}{\textbf{Method}}  & 
\multicolumn{2}{c}{\textbf{APTv2}} \\
\cmidrule(lr){2-3}
 &  \textbf{PCK@0.1$_{\text{img}}$} & \textbf{PCK@0.05$_{\text{img}}$} \\
\hline 
AllTracker~\cite{harley2025alltracker} & 88.6 & 74.5\\
CoTracker3~\cite{karaev2025cotracker3} & 87.1 & 74.5\\
TAPIR~\cite{doersch2023tapir} & 84.1 & 70.1\\
BootsTAPIR v2~\cite{doersch2024bootstap} & 85.6 & 72.3\\
Ours (DHf w/) & 86.6 & 70.5\\
\bottomrule
\end{tabular}
}
\captionof{table}{Comparison with representative Track Any Point methods on APTv2.}
\label{tab_unsuper_compare}
\end{minipage}
\hfill
\begin{minipage}[t]{0.55\textwidth}
\vspace{-0.5em}
\centering
\begin{tikzpicture}
  \tikzset{
    rotated_label/.style={rotate=90, font=\tiny, anchor=center, inner sep=1pt},
    header_label/.style={font=\tiny\bfseries, anchor=center, inner sep=2pt},
    image_node/.style={anchor=center, inner sep=0pt, outer sep=0pt}
  }

  \newcommand{\bufOne}{\includegraphics[trim=1.5cm 2.5cm 1.5cm 3.5cm, clip, height=1cm, width=0.47\linewidth]{"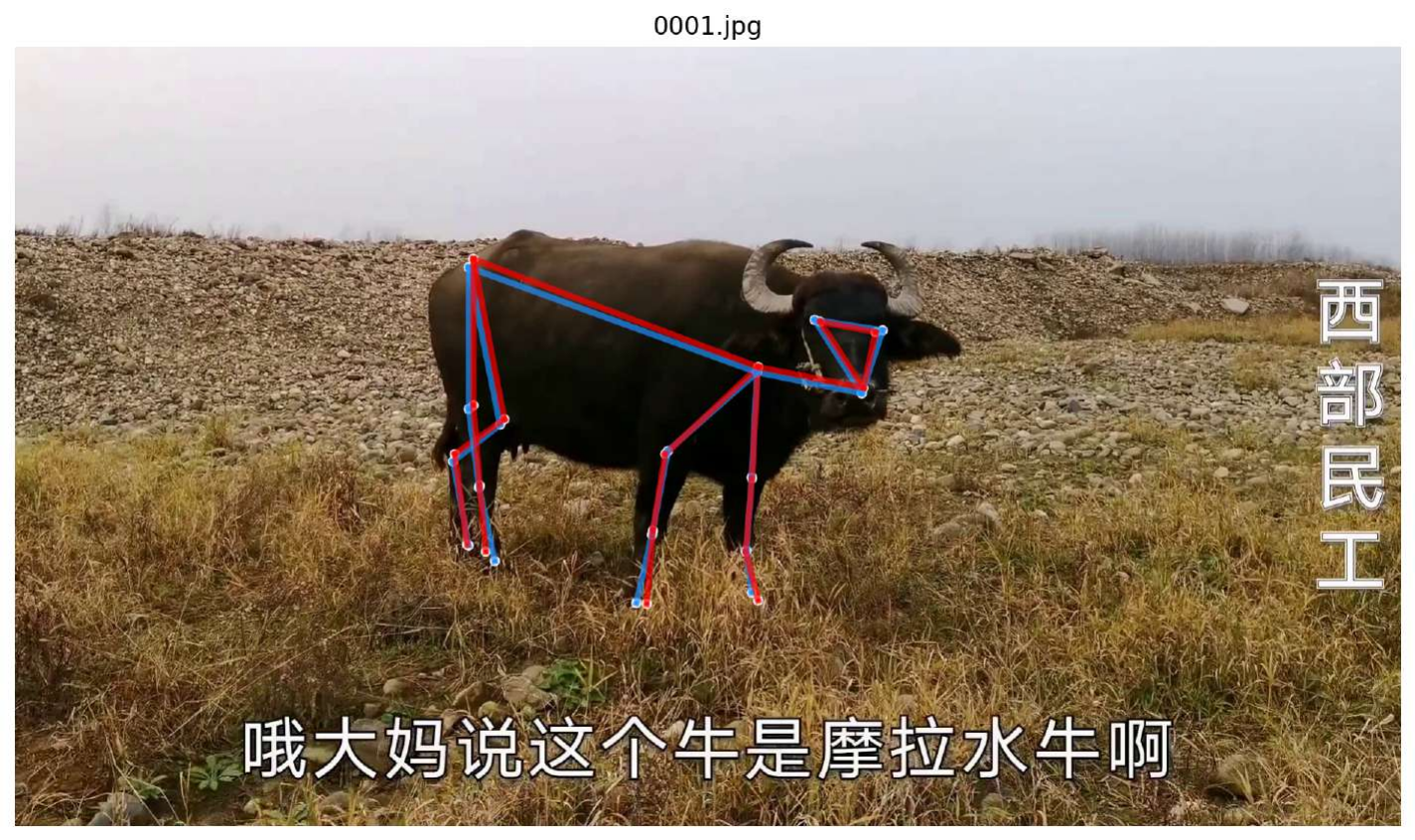"}}
  \newcommand{\monOne}{\includegraphics[trim=1.5cm 2cm 1.5cm 4cm, clip, height=1cm, width=0.47\linewidth]{"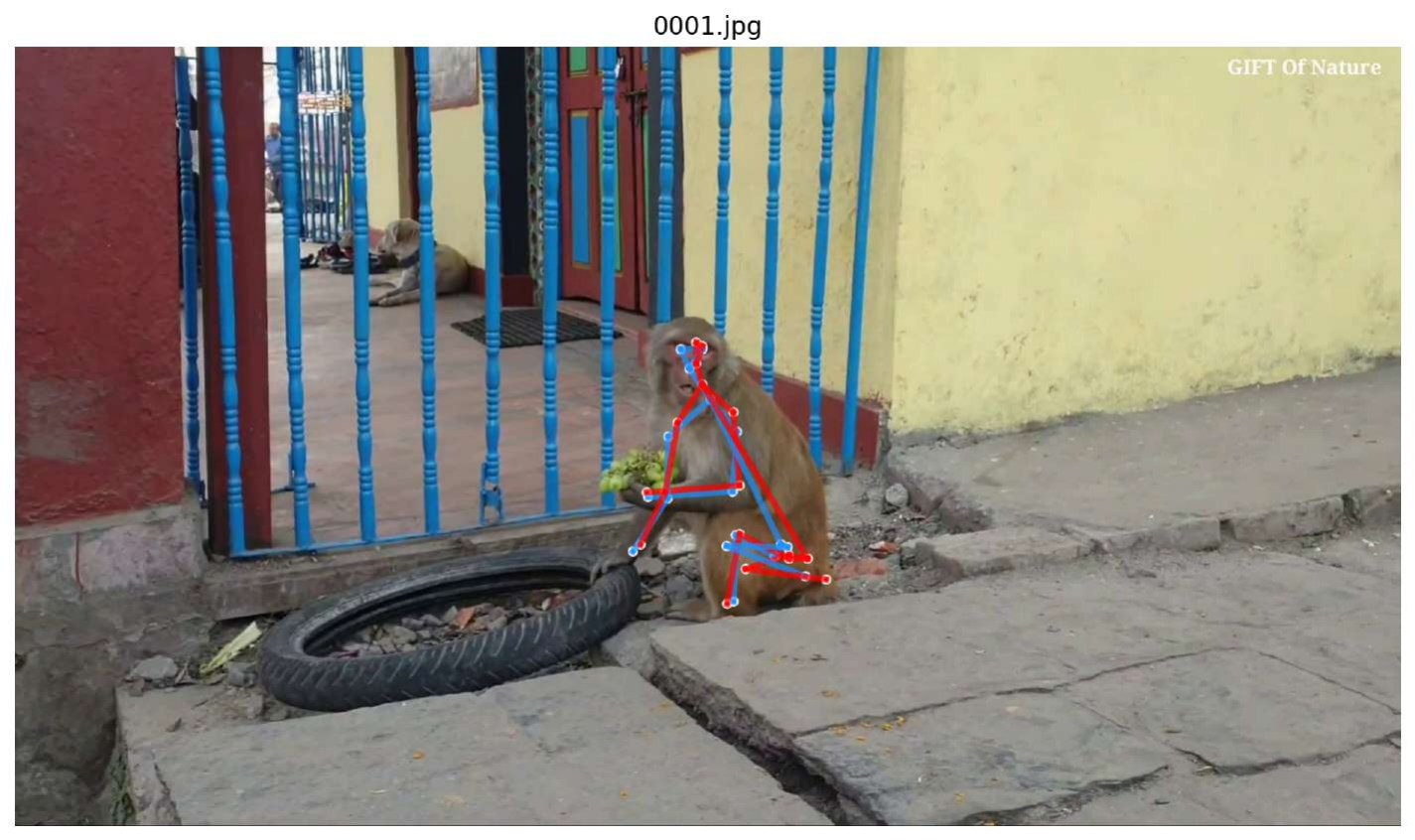"}}
  
  \newcommand{\bufSeven}{\includegraphics[trim=1.5cm 2.5cm 1.5cm 3.5cm, clip, height=1cm, width=0.47\linewidth]{"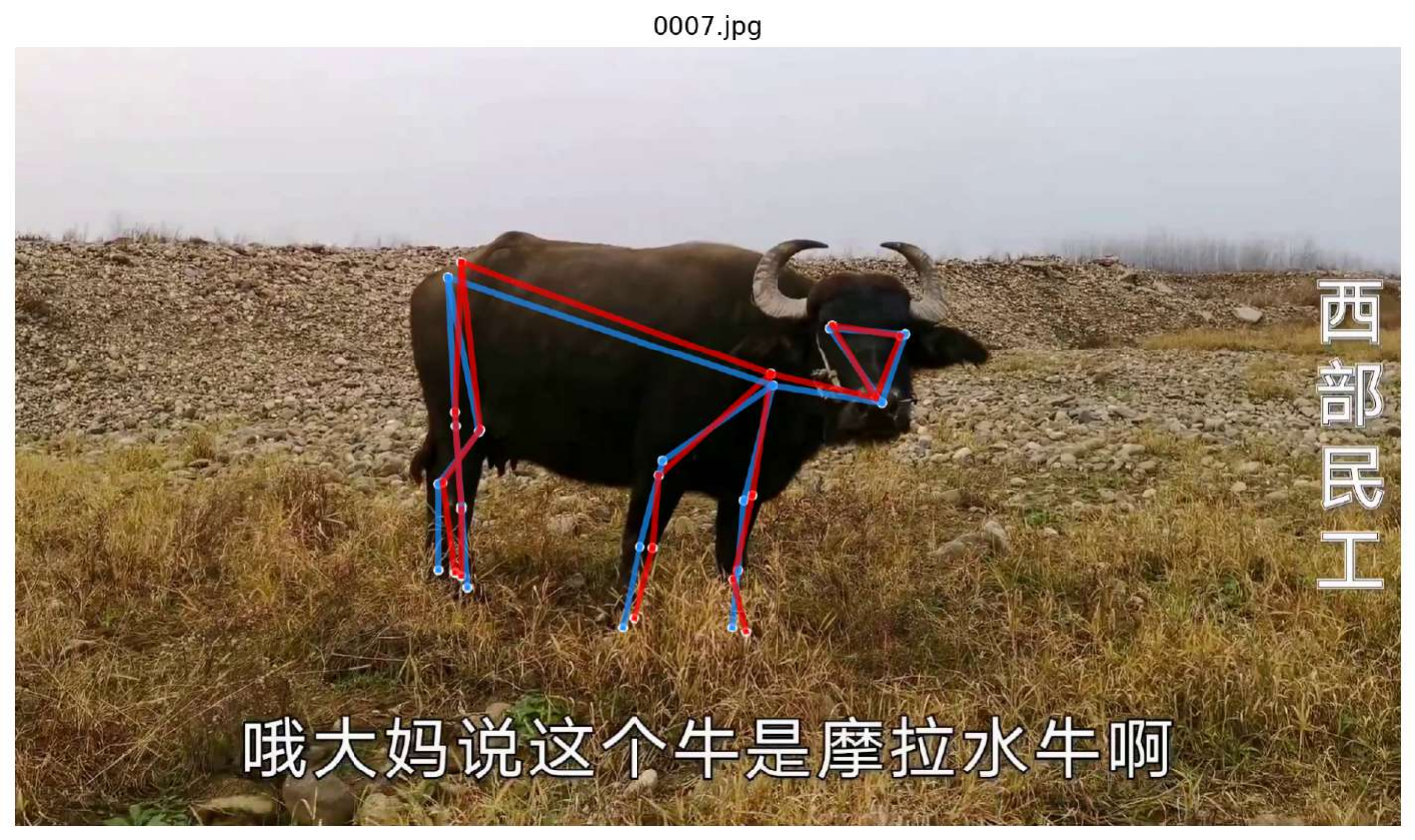"}}
  \newcommand{\monSeven}{\includegraphics[trim=1.5cm 2cm 1.5cm 4cm, clip, height=1cm, width=0.47\linewidth]{"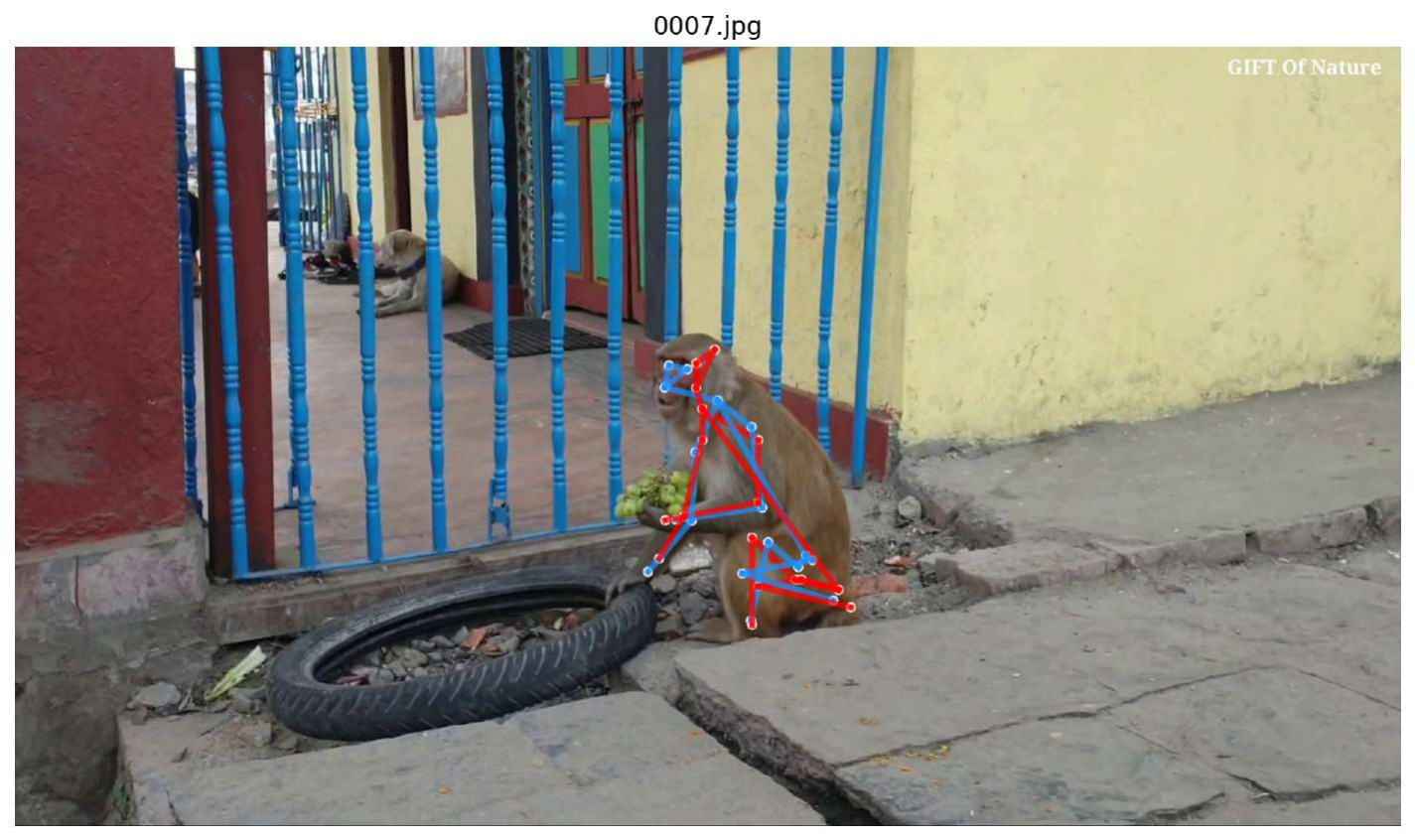"}}
  
  \newcommand{\bufFourteen}{\includegraphics[trim=1.5cm 2.5cm 1.5cm 3.5cm, clip, height=1cm, width=0.47\linewidth]{"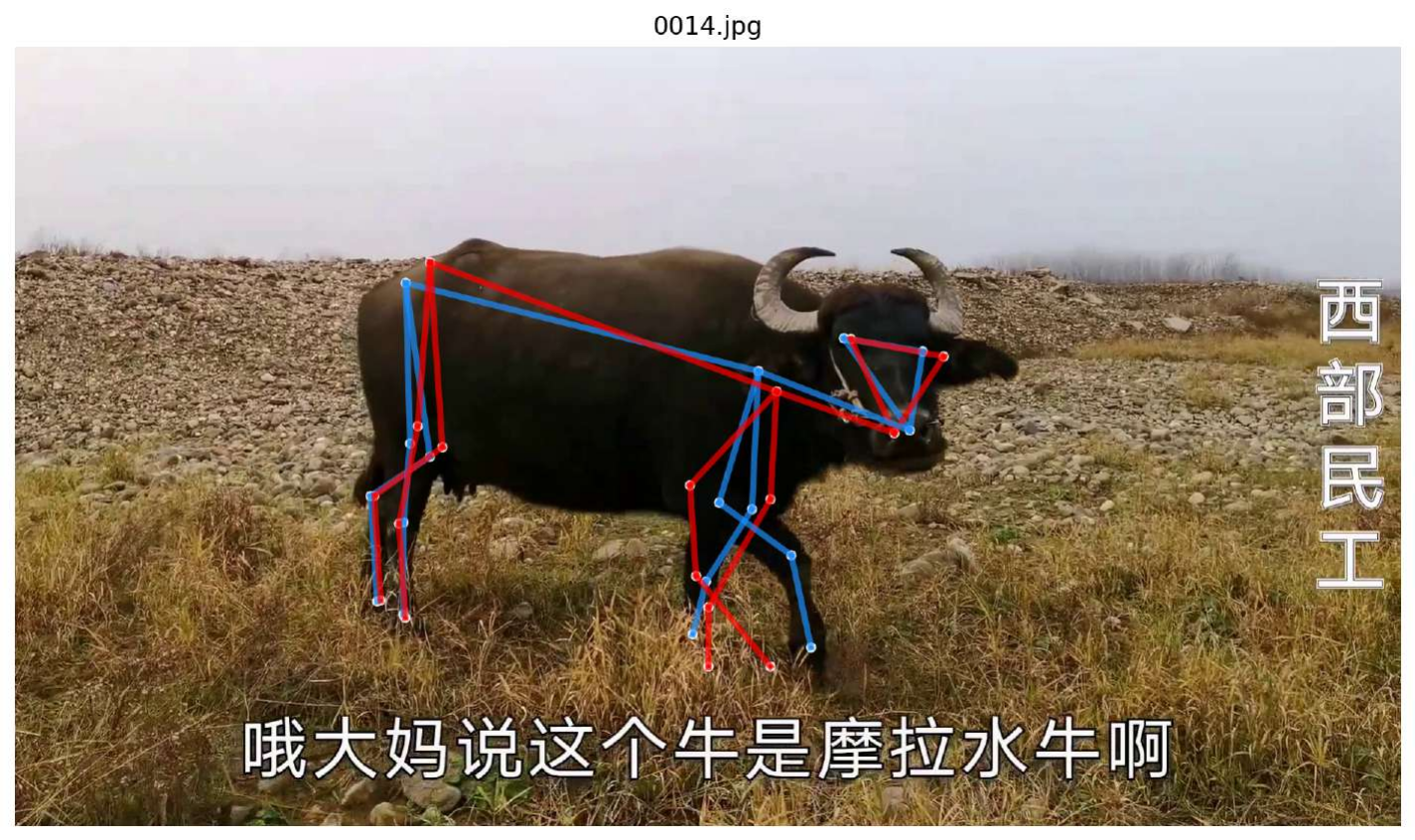"}}
  \newcommand{\monFourteen}{\includegraphics[trim=1.5cm 2cm 1.5cm 4cm, clip, height=1cm, width=0.47\linewidth]{"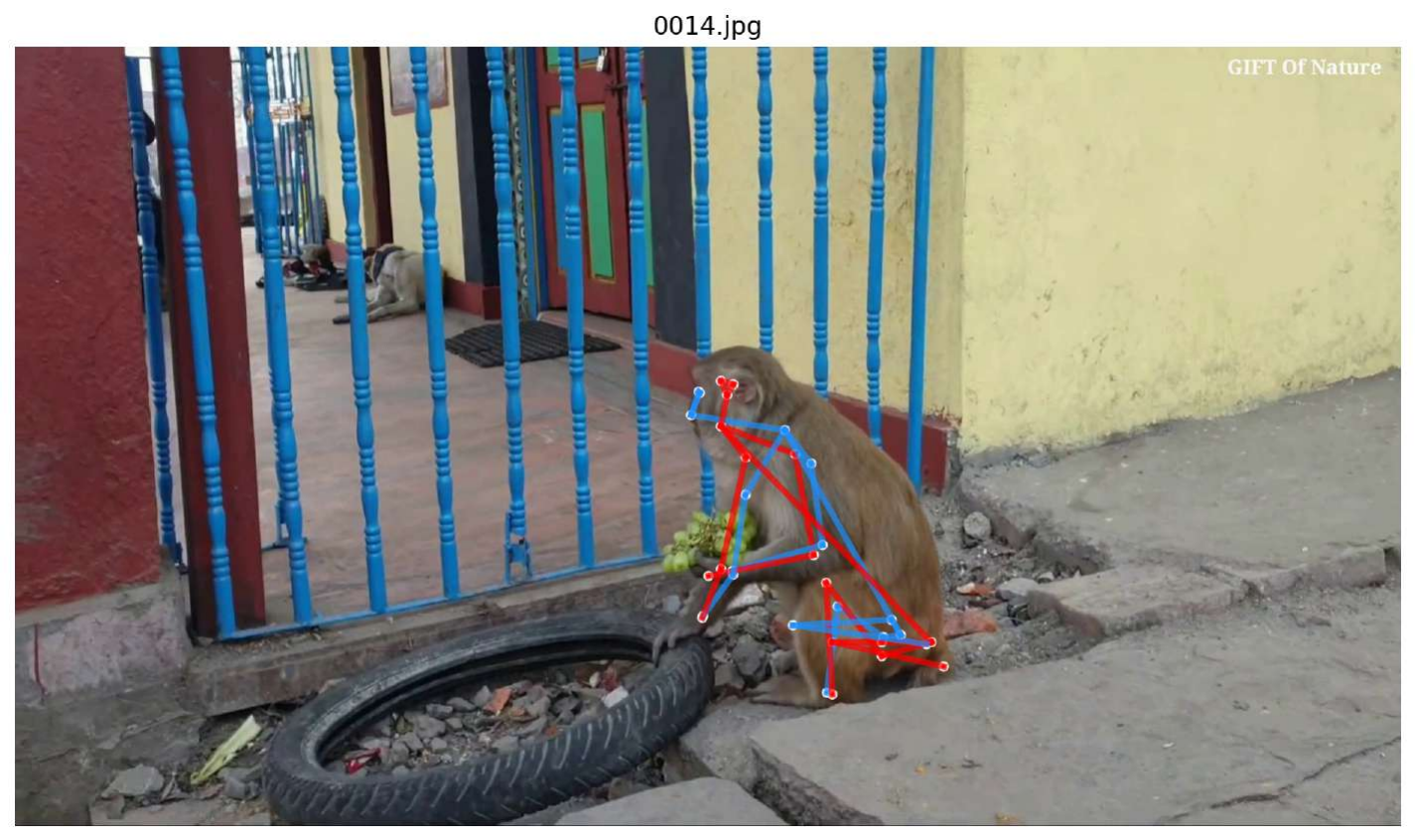"}}

  \matrix (qualgrid) [
    matrix of nodes,
    ampersand replacement=\&, 
    row sep=2pt,      
    column sep=2pt,   
    nodes={image_node}
  ] {
    \& |[header_label]| Buffalo v9C2 \& |[header_label]| Monkey v6C5 \\
    |[rotated_label]| Fr. 1 \& \bufOne \& \monOne \\
    |[rotated_label]| Fr. 7 \& \bufSeven \& \monSeven \\
    |[rotated_label]| Fr. 14 \& \bufFourteen \& \monFourteen \\
  };

\end{tikzpicture}
\vspace{-2em}
\captionof{figure}{Comparison of tracking results across increasing temporal distances. Ground truth is shown in blue and predictions in red.}
\label{fig:qualitative}
\end{minipage}

\end{figure*}

Table~\ref{tab_unsuper_compare} compares our unsupervised, foundation-model correspondence approach with state-of-the-art Track Any Point (TAP) methods. The comparison is deliberately asymmetric. Existing TAP methods are trained explicitly for point tracking on large-scale tracking and optical-flow datasets --- several of which contain animal subjects (TAP-Vid~\cite{doersch2022tap}, DynamicReplica~\cite{karaev2023dynamicstereo}, PointOdyssey~\cite{zheng2023pointodyssey}) --- and exploit temporal information across the entire video. Ours performs pairwise frame matching on generic vision foundation-model features, with no tracking-specific training and no temporal context.
Despite these disadvantages, our method performs competitively with the latest dedicated trackers. This suggests that generic foundation-model features already capture most of the correspondence capability point tracking requires, and that a simple matching pipeline can approach large end-to-end trackers without learning the task. The pairwise design also operates without lookahead, making it directly applicable to online or streaming settings such as real-time animal monitoring.




\subsubsection{Fine-grained Analysis}
Table \ref{tab_unsuper_bodyregion} breaks down performance across different body regions. We bin the 17 keypoints in APTv2 into three groups: head, central body, and limb extremities. We report the average PCK@0.05$_{\text{img}}$ for these regions, while per-keypoint results are provided in Table \ref{taba2} of the Appendix.

\begin{table*}[ht!]
\centering
\resizebox{0.9\textwidth}{!}{
\begin{tabular}{lcccccc}
\toprule
\textbf{Body region} & \textbf{DINOv3-Last} & \textbf{DINOv3-All} & \textbf{CD w/o} & \textbf{CD w/} & \textbf{DHf w/o} & \textbf{DHf w/} \\
\hline 

Head
 & 51.8 & 53.1 & 69.2 & 71.3 & 85.7 & \underline{88.2} \\
\addlinespace

Central body
 & 49.3 & 51.1 & 71.4 & 72.4 & 75.5 & \underline{76.7} \\
\addlinespace

Limb extremities
 & 46.0 & 47.3 & 60.1 & 60.8 & 65.8 & \underline{67.2} \\
\bottomrule
\end{tabular}}
\caption{Unsupervised results on APT-v2 across body regions. The head, central body, and limb extremities contain 3, 6, and 8 keypoints, respectively. Values report mean PCK@0.05$_{\text{img}}$ per region. Best-performing backbones are \underline{underlined}.}
\label{tab_unsuper_bodyregion}
\end{table*}
Performance varies systematically with body region. Four of the six backbones achieve their best results on the head and the remaining two on the central body --- regions where keypoints are anatomically stable and undergo comparatively simple motion. Limb extremities are consistently the hardest: for the CleanDIFT- and Diffusion Hyperfeatures-based backbones, PCK@0.05$_{\text{img}}$ on distal keypoints falls more than 10\%, and in some cases over 20\%, below each backbone's best region. Fast motion, self-occlusion, and appearance ambiguity between left and right limbs all compound here, and the gap indicates that fine-grained articulated motion in distal parts remains unsolved by current foundation-model features.

We further report unsupervised results on APT-v2 in terms of PCK@0.05$_{\text{img}}$ for individual species. Table \ref{tab_unsuper_families} presents the three best-performing and three worst-performing species among the 30 species in APT-v2. Results for the remaining species are provided in Table \ref{taba3} of the Appendix.

\begin{table*}[ht!]
\centering
\resizebox{\textwidth}{!}{
\begin{tabular}{llcccccc}
\toprule
\textbf{Family} & \textbf{Species (Number)} & \textbf{DINOv3-Last} & \textbf{DINOv3-All} & \textbf{CD w/o} & \textbf{CD w/} & \textbf{DHf w/o} & \textbf{DHf w/} \\
\hline

Giraffidae & Giraffe (18)
& 53.4 & 56.3 ($\uparrow$2.9)
& 76.7 & 77.8 ($\uparrow$1.1)
& 80.7 & 83.1 ($\uparrow$2.4) \\
\addlinespace

Equidae & Zebra (13)
& 55.9 & 58.0 ($\uparrow$2.1)
& 75.1 & 76.0 ($\uparrow$0.9)
& 81.4 & 82.4 ($\uparrow$1.0) \\
\addlinespace

Bovidae & Cow (28)
& 52.5 & 54.6 ($\uparrow$2.1)
& 73.5 & 73.8 ($\uparrow$0.3)
& 79.3 & 79.8 ($\uparrow$0.5) \\
\addlinespace

Canidae & Fox (26)
& 34.8 & 35.7 ($\uparrow$0.9)
& 50.3 & 51.9 ($\uparrow$1.6)
& 61.6 & 62.2 ($\uparrow$0.6) \\
\addlinespace

Hominidae & Chimpanzee (7)
& 33.7 & 33.7 ($\uparrow$0.0)
& 47.9 & 50.6 ($\uparrow$2.7)
& 58.0 & 61.1 ($\uparrow$3.1) \\
\addlinespace

Hominidae & Gorilla (10)
& 27.7 & 27.2 ($\downarrow$0.5)
& 48.5 & 50.7 ($\uparrow$2.2)
& 56.5 & 60.3 ($\uparrow$3.8) \\
\midrule

\multicolumn{2}{l}{Std. across species}
& 7.19
& 8.21
& 8.34
& 7.99
& 6.81
& 6.30 \\
\bottomrule
\end{tabular}}
\caption{Unsupervised results on APT-v2 across selected families. Values report mean PCK@0.05$_{\text{img}}$ per family. $\uparrow$ indicates performance change relative to the previous column. Std. across species denotes the standard deviation over all 30 APT-v2 species.}
\label{tab_unsuper_families}
\end{table*}

The top-performing species (Giraffe, Zebra, and Cow) exhibit distinctive body structures and clear visual textures, typically appearing in relatively uncluttered grassland scenes, which facilitates reliable feature matching and keypoint localisation. In contrast, the lowest-performing species (Fox, Chimpanzee, and Gorilla) present more challenging conditions, including small object scale, appearance ambiguity, and frequent occlusions in cluttered environments such as zoo settings. These factors make cross-instance and fine-grained correspondence more difficult under large pose variations. The relatively high standard deviation across species further reflects the variability in performance under different visual and structural conditions. Figure~\ref{fig_cases} illustrates successful and failure cases of the unsupervised route. The use of prompted, animal-specific diffusion features improves performance across the board, for example: Giraffe (+2.4), Chimpanzee (+3.1) and Gorilla (+3.8).  

To further demonstrate the flexibility of our framework, we additionally evaluate tracking performance on user-defined intermediate points. The corresponding results are provided in Table \ref{taba4} of the Appendix.

\subsubsection{Ablation Study}

\begin{table*}[ht!]
\centering
\resizebox{\textwidth}{!}{
\begin{tabular}{llcccc}
\toprule
\textbf{Backbone} & \textbf{Implementation} & \begin{tabular}[c]{@{}c@{}}\textbf{PCK@0.1$_{\text{img}}$} \\ \textbf{(w/o corr.)}\end{tabular} & \textbf{PCK@0.1$_{\text{img}}$} & \begin{tabular}[c]{@{}c@{}}\textbf{PCK@0.05$_{\text{img}}$} \\ \textbf{(w/o corr.)}\end{tabular} & \textbf{PCK@0.05$_{\text{img}}$} \\
\hline

BioClip & Last-4-layer concatenation
& 28.5 & 42.9 ($\uparrow$14.4)
& 15.5 & 21.9 ($\uparrow$6.4) \\
\addlinespace

DINOv3 & Last-layer patch features
& 64.6 & 67.9 ($\uparrow$3.3)
& 40.4 & 42.4 ($\uparrow$2.0) \\
& All-layer concatenation
& 66.6 & 69.4 ($\uparrow$2.8)
& 42.1 & 43.9 ($\uparrow$1.8) \\
\addlinespace

CleanDIFT & w/o prompt on species type
& 80.3 & 82.6 ($\uparrow$2.3)
& 59.0 & 60.6 ($\uparrow$1.6) \\
& w prompt on species type
& 82.2 & 84.0 ($\uparrow$1.8)
& 60.6 & 61.9 ($\uparrow$1.3) \\
\addlinespace

\begin{tabular}[t]{@{}l@{}}DHf\end{tabular} & w/o prompt on species type
& 83.8 & 85.1 ($\uparrow$1.3)
& 67.7 & 68.6 ($\uparrow$0.9) \\
& w prompt on species type
& 85.6 & 86.6 ($\uparrow$1.0)
& 69.9 & 70.5 ($\uparrow$0.6) \\
\bottomrule
\end{tabular}}
\caption{Ablation study of drift correction in the unsupervised route. $\uparrow$ denotes performance change relative to the previous column.}
\label{tabconfused}
\end{table*}

The ablation over our drift correction module in Table~\ref{tabconfused} demonstrates that it consistently improves all unsupervised backbones by reducing accumulated matching errors during tracking. Qualitative visualizations are in Figure \ref{figa1} of the Appendix.

\subsubsection{Temporal Robustness Analysis}

Since APT-v2 consists of fixed-length clips of 15 frames, we further analyse PCK@0.1$_{\text{img}}$ and PCK@0.05$_{\text{img}}$ with respect to frame index. As shown in Figure \ref{fig:qualitative}, performance gradually decreases as the temporal distance from the reference frame ($r=0$) increases. Since each target frame is matched directly to the reference frame rather than propagated sequentially, this degradation is mainly caused by increasing appearance and pose variations. Figure \ref{figa2} in the Appendix further shows that selecting the reference frame from the middle of the sequence results in less pronounced degradation in both temporal directions. 

\subsection{Results on the supervised route}
\label{sec4_3}
In this section, we present detailed experimental results for the supervised route. We evaluate the overall performance of our method and further assess its generalizability under a leave-one-out setting across six animal families.

\subsubsection{Overall Performance}

\begin{table}[ht!]
\centering
\resizebox{0.85\linewidth}{!}{
\begin{tabular}{llcc}
\toprule
\textbf{Method} & \textbf{Implementation} & \textbf{PCK@0.1$_{\text{img}}$} & \textbf{PCK@0.05$_{\text{img}}$} \\
\hline

SimpleBaseline~\cite{yang2023aptv2} 
 & ResNet50 & 98.2 & 94.1 \\
 & ResNet101 & 98.0 & 94.0 \\
\addlinespace

\multirow{2}{*}{HRFormer~\cite{yang2023aptv2}} 
 & Small & 98.2 & 94.4 \\
 & Base & 98.3 & 94.5 \\
\addlinespace

\multirow{2}{*}{HRNet~\cite{yang2023aptv2}} 
 & W32 & 98.4 & 94.8 \\
 & W48 & 98.5 & \underline{95.2} \\
\midrule

Ours (DINOv3)
 & Concat - 2 layers & 91.8 & 74.0 \\
\addlinespace

\begin{tabular}[t]{@{}l@{}}Ours (DHf)\end{tabular}
 & w/o prompt on species type & \underline{99.1} & 94.7 \\
\bottomrule
\end{tabular}}
\caption{Supervised results on APT-v2 across different backbones. The best results are \underline{underlined}.}
\label{tab_apt_super}
\end{table}














Table \ref{tab_apt_super} compares supervised performance across backbones on APTv2. By incorporating reference-frame keypoints via a keypoint encoder, our method achieves strong performance against human pose estimation baselines. For a fair comparison, all baselines~\cite{yang2023aptv2} are re-evaluated using our PCK protocol.

It is worth noting that these baselines are first pre-trained on large-scale human pose datasets such as Microsoft COCO~\cite{lin2014microsoft} for 200 epochs before being fine-tuned on APTv2. In contrast, our backbone remains frozen, and only the keypoint prompt encoder, feature projection module, and correspondence matcher are trained for 50 epochs. Despite this lightweight adaptation strategy, the proposed framework achieves strong performance, demonstrating that frozen foundation-model features can be effectively adapted for animal pose tracking.

\subsubsection{Cross-Family Generalization}

\begin{table*}[htbp]
\centering

\resizebox{0.95\textwidth}{!}{
\begin{tabular}{l ccccccc}
\toprule
\textbf{Training $\setminus$ Testing}
&\textbf{ Canidae }& \textbf{Felidae} & \textbf{Hominidae} & \textbf{Cercopithecidae} & \textbf{Ursidae} & \textbf{Bovidae} \\
\midrule

w/o Canidae         & \cellcolor{gray!25}51.4 $_{\scriptsize 67.1}$ & 96.9 & 95.3 & 95.9 & 97.6 & 97.2 \\
w/o Felidae         & 80.7 & \cellcolor{gray!25}55.8 $_{\scriptsize 70.4}$ & 94.1 & 95.1 & 96.9 & 96.6 \\
w/o Hominidae       & 80.4 & 96.3 & \cellcolor{gray!25}41.9 $_{\scriptsize 61.6}$ & 94.8 & 96.8 & 96.2 \\
w/o Cercopithecidae & 80.9 & 96.1 & 94.2 & \cellcolor{gray!25}45.5 $_{\scriptsize 65.2}$  & 96.6 & 96.0 \\
w/o Ursidae         & 82.1 & 96.4 & 94.4 & 95.2 & \cellcolor{gray!25}54.9 $_{\scriptsize 68.8}$ & 96.3 \\
w/o Bovidae         & 77.7 & 95.2 & 93.5 & 94.5 & 95.9 & \cellcolor{gray!25}55.8 $_{\scriptsize 76.6}$\\

\midrule

Average (seen)      & 80.4 & 96.2 & 94.3 & 95.1 & 96.8 & 96.5 \\
$\Delta$ (seen-unseen) & 29.0 & 40.4 & 52.4 & 49.6 & 41.9 & 40.7 \\

\bottomrule
\end{tabular}
}

\caption{PCK@0.05$_{\text{img}}$ under the leave-one-out setting on APTv2. Each diagonal gray cell denotes the unseen test category excluded during training. The small values annotated on the diagonal correspond to the performance of the unsupervised route for the same category. $\Delta$ represents the gap between seen and unseen performance.}
\label{tab:leave_one_out}

\end{table*}

Since performance across animal families directly reflects cross-category generalization, we further conduct a leave-one-out experiment over six animal families in the supervised setting. For each family, all corresponding training samples are removed and the model is evaluated only on the held-out family. As shown in Table \ref{tab:leave_one_out}, performance on unseen families drops substantially compared with seen categories. Moreover, the leave-one-out results are often lower than those achieved by the unsupervised route on the same families (diagonal values), suggesting that supervised optimization tends to specialize to the training distribution and generalizes less effectively to novel categories. In contrast, the unsupervised route exhibits more stable cross-family transfer. A qualitative illustration is provided in Figure \ref{figa3} of the Appendix.

\subsubsection{Comparison Between Supervised and Unsupervised Routes}
\begin{figure}[!h]
	\centering
	\includegraphics[width=\textwidth]{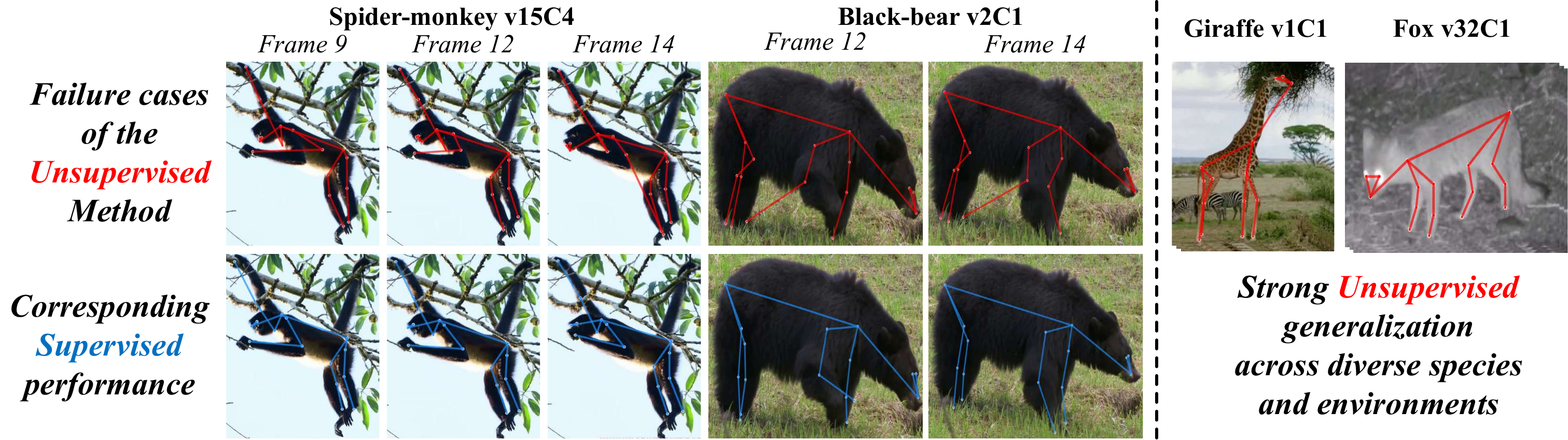}
	\caption{
Comparison between supervised and unsupervised routes. Left: failure cases of the unsupervised method where supervised learning yields more accurate predictions under challenging conditions. Right: successful unsupervised results across diverse species and environments, highlighting strong cross-domain generalization.
}
	\label{fig_cases}
\end{figure}

Figure \ref{fig_cases} provides qualitative evidence of this complementary behaviour. While supervised learning produces more accurate predictions in challenging cases involving occlusion, ambiguity, or large appearance changes, the unsupervised route demonstrates stronger robustness across diverse species and environments. Together, these results indicate that supervised and unsupervised routes offer complementary advantages in animal pose tracking.

\section{Conclusion}
\label{sec5}
We present a promptable animal pose tracking framework for tracking user-specified keypoints across video sequences. It comprises supervised and unsupervised routes, enabling accurate tracking with and without manual annotations, respectively.

Built upon vision foundation model representations, our framework provides a flexible solution for animal pose tracking across diverse real-world scenarios. Extensive experiments on APTv2 and TigDog demonstrate that the supervised route achieves strong tracking accuracy across multiple species and challenging environments, while the unsupervised route delivers competitive performance without requiring any annotated training data.

The results demonstrate that dense foundation-model representations offer an effective basis for animal keypoint correspondence across diverse species and environments. Moreover, our analyses reveal a clear trade-off between the two paradigms. The supervised route excels in challenging cases involving occlusion, ambiguity, and large appearance changes, whereas the unsupervised route exhibits stronger robustness to category shifts and unseen species. These findings suggest that supervised and unsupervised routes provide complementary strengths for generalized animal pose tracking.

Finally, although the proposed unsupervised route already demonstrates strong cross-species generalization, its performance may be further improved through more advanced correspondence estimation and matching strategies, which we leave for future work.

\section*{Acknowledgments}
Daniela Ivanova and Nicolas Pugeault acknowledge support from the NERC project DEAL – DEcentrAlised Learning for automated image analysis and biodiversity monitoring (ref: UKRI041), and Le Li is supported by a UKRI EPSRC Scholarship.

\bibliographystyle{unsrt}  
\bibliography{references}  
\clearpage

\appendix

\renewcommand{\thefigure}{A\arabic{figure}}
\renewcommand{\thetable}{A\arabic{table}}

\setcounter{figure}{0}
\setcounter{table}{0}

\begin{center}
{\large\bfseries
Promptable Animal Pose Tracking Across Species
}

\vspace{1.2em}

\large{Appendix}

\end{center}

\vspace{1em}

\begin{center}
\begin{minipage}{0.9\textwidth}
\centering
This appendix provides supplementary material for
\textit{Promptable Animal Pose Tracking Across Species},
including further comparisons with existing methods,
implementation details, additional experimental results,
and qualitative visualizations.
\end{minipage}
\end{center}

\vspace{1.5em}

\section{Additional Comparisons with Animal Tracking Methods}

This section provides comparisons with classic animal pose estimation and tracking methods for Section \ref{sec:intro} in the main paper. Compared with state-of-the-art approaches, our method introduces an innovative tracking architecture together with a fully unsupervised alternative route.

\begin{table*}[htbp]
\centering

\small
\setlength{\tabcolsep}{3pt} 

\resizebox{\textwidth}{!}{
\begin{tabular}{l p{0.22\textwidth} p{0.22\textwidth} p{0.22\textwidth} p{0.22\textwidth}}
\toprule

\textbf{Method}
& \centering \includegraphics[width=\linewidth, keepaspectratio]{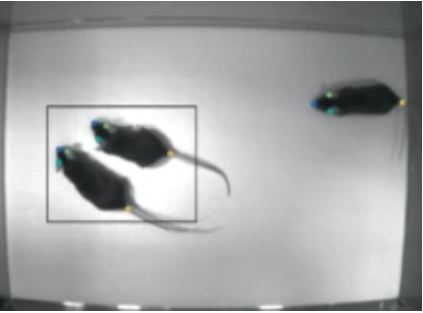}\par DeepLabCut (2018 Nature)~\cite{mathis2018deeplabcut}
& \centering \includegraphics[width=\linewidth, keepaspectratio]{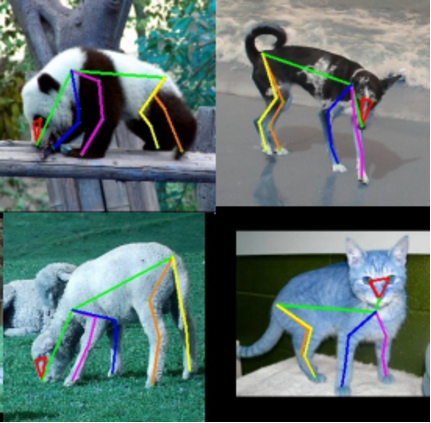}\par ScarceNet (2023 CVPR)~\cite{li2023scarcenet}
& \centering \includegraphics[width=\linewidth, keepaspectratio]{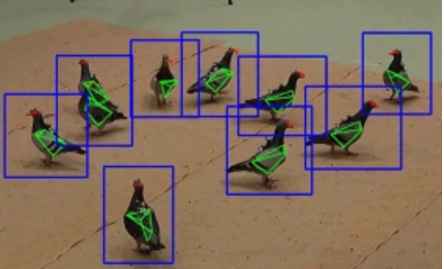}\par 3D-MuPPET (2024 IJCV)~\cite{waldmann20243d}
& \centering \includegraphics[width=\linewidth, keepaspectratio]{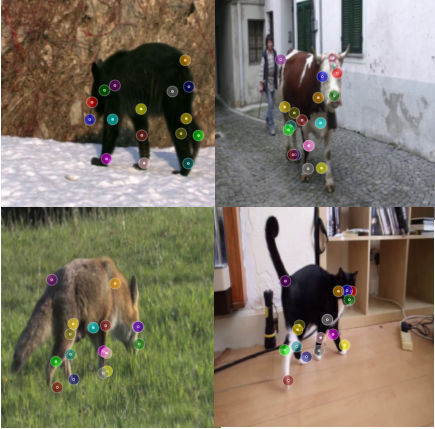}\par \textbf{Ours: Promptable Animal Pose Tracking}
\tabularnewline

\midrule

\textbf{Supervision}
& \centering Supervised
& \centering Pseudo-label-based supervised
& \centering Supervised
& \centering Supervised / Unsupervised
\tabularnewline
\addlinespace

\textbf{Dataset}
& \centering Odor trail tracking dataset
& \centering AP-10K and TigDog dataset
& \centering 3D-POP dataset
& \centering APTv2 and TigDog dataset
\tabularnewline
\addlinespace

\textbf{Species (\#)}
& \centering 1
& \centering 50
& \centering 1
& \centering 30
\tabularnewline
\addlinespace

\textbf{Backbone}
& \centering ResNet-based CNN
& \centering Multi-animal CNN network
& \centering Multi-view CNN network
& \centering Foundation models
\tabularnewline
\addlinespace

\textbf{Scope}
& \centering Laboratory environments
& \centering Diverse scenarios
& \centering 3D multi-animal tracking
& \centering Diverse real-world scenarios
\tabularnewline

\bottomrule
\end{tabular}
}
\vspace{0.1cm}
\caption{Comparison of representative animal pose estimation and tracking methods.}
\label{taba1}

\end{table*}

\newpage
\section{Datasets and Implementation Details}
\label{sec:supp_datasets_impl}
In this section, we provide additional details on the evaluation datasets and implementation described in Section \ref{sec4} of the main paper.

In terms of the \textbf{datasets}, we provide a more detailed description of the APTv2 dataset. As mentioned in the main paper, APTv2 contains 30 animal species across 15 families, with each video consisting of 15 frames. The 15 frames were not uniformly sampled at a fixed frame rate. Instead, they were manually selected to ensure that the animal exhibited noticeable motion between sampled frames. 

In our setting, keypoints annotated in a single reference frame are propagated to the remaining 14 frames for evaluation. All experiments reported in this paper use the first frame as the reference frame for consistency. However, the proposed method is not restricted to this choice and can use any frame as the reference frame. In practice, using a reference frame from the middle of the sequence often leads to better performance. 

The dataset is divided into Easy and Hard subsets, and we focus on the Easy subset where each video contains a single target instance. However, multiple animals may still appear in the same scene, leading to inter-instance ambiguity and distracting objects, and the Easy subset still contains complex scenes and occlusions. 

Following human pose benchmarks such as Microsoft COCO~\cite{lin2014microsoft}, APTv2 provides COCO-style instance annotations and bounding boxes. However, we do not rely on cross-frame identity annotations or instance tracking labels during evaluation. Instead, instance association is inferred from spatial cues derived from bounding boxes and predicted keypoint locations, which aligns with our promptable setting where tracking is initialized from a single reference frame.

In terms of the \textbf{implementation details}, we use the pre-trained aggregation\_network.pt released with Diffusion Hyperfeatures \cite{luo2023diffusion} for our best-performing backbone, Diffusion Hyperfeatures (DHf). The input image resolution is set to 224 $\times$ 224. The channels of the Keypoint Prompt Encoder $K$ is set as 17 for APTv2 and 19 for TigDog.

\newpage
\section{Additional Results for the Unsupervised Route}

Table \ref{taba2} provides the full per-region breakdown corresponding to the summary reported in the main paper. We additionally report results for each backbone under the three keypoint groups: head, central body, and limb extremities. This extended view allows finer-grained comparison of representation behavior across different anatomical regions.

\begin{table}[htbp]
\centering
\label{tabconfused1}

\resizebox{\textwidth}{!}{
\begin{tabular}{l l c c c c c c}
\toprule
\textbf{Body region} & \textbf{Joint (Number)} & \textbf{DINOv3-Last} & \textbf{DINOv3-All} & \textbf{CD w/o} & \textbf{ CD w/} & \textbf{DHf w/o} & \textbf{DHf w/} \\
\midrule

Head & Left eye (1) &52.4 &53.7 &70.8 &73.1 &86.7 &89.2 \\
& Right eye (2) &52.8 &54.0 &70.4 &72.8 &85.9 &88.5 \\
& Nose (3) &50.2 &51.7 &66.3 &67.9 &84.6 &86.8 \\
& Average &\textbf{51.8} &\textbf{53.1} &\textbf{69.2} &\textbf{71.3} &\textbf{85.7} &\textbf{88.2} \\
\addlinespace

Central body & Neck (4) &48.7 &49.8 &72.2 &73.2 &76.7 &78.0 \\
& Tail (5) &55.3 &57.1 &73.4 &74.0 &80.2 &80.6 \\
& Left shoulder (6) &46.3 &47.7 &70.3 &71.8 &73.7 &75.4 \\
& Right shoulder (9) &46.2 &47.9 &69.0 &70.4 &72.9 &74.5 \\
& Left hip (12) &49.2 &51.9 &71.8 &72.6 &74.3 &75.9 \\
& Right hip (15) &50.0 &52.3 &71.6 &72.6 &74.9 &76.0 \\
& Average &\textbf{49.3} &\textbf{51.1} &\textbf{71.4} &\textbf{72.4} &\textbf{75.5} &\textbf{76.7} \\
\addlinespace

Limb extremities & Left elbow (7) &42.9 &44.1 &60.8 &62.1 &66.0 &68.5 \\
& Right elbow (10) &42.8 &43.8 &61.0 &61.8 &66.3 &68.2 \\
& Left knee (13) &46.9 &49.4 &63.5 &64.5 &68.8 &70.1 \\
& Right knee (16) &46.6 &48.8 &63.8 &65.0 &68.7 &70.9 \\
& Left hand (8) &44.3 &45.0 &55.6 &56.0 &61.8 &62.7 \\
& Right hand (11) &44.4 &45.0 &54.7 &55.2 &61.5 &62.4 \\
& Left foot (14) &49.5 &50.4 &59.6 &60.4 &65.6 &66.6 \\
& Right foot (17) &50.2 &52.2 &61.6 &61.6 &67.3 &68.3 \\
& Average &\textbf{46.0} &\textbf{47.3} &\textbf{60.1} &\textbf{60.8} &\textbf{65.8} &\textbf{67.2} \\
\bottomrule
\end{tabular}
}

\caption{\centering Unsupervised results on APT-v2 across body regions. CD denotes CleanDIFT, and DHf denotes Diffusion Hyperfeatures. Values report mean PCK@0.05$_{\text{img}}$ per region. Bold indicates the average.}
\label{taba2}
\end{table}
\newpage
Table \ref{taba3} reports additional unsupervised results across animal families, complementing the species-level analysis in the main paper. We highlight representative high-performing and low-performing categories among the 30 species in the main paper, while the complete results are included in appendix.

\begin{table}[htbp]
\centering
\label{tabconfused1}

\resizebox{\textwidth}{!}{
\begin{tabular}{l l c c c c c c}
\toprule
\textbf{Family} & \textbf{Species (Number)} & \textbf{DINOv3-Last} & \textbf{DINOv3-All} & \textbf{CD w/o} & \textbf{CD w/} & \textbf{DHf w/o} & \textbf{DHf w/} \\
\midrule

Cervidae & Deer (1) &40.2 &43.6 ($\uparrow$3.4) &62.2 &63.0 ($\uparrow$0.8) &71.3 &72.8 ($\uparrow$1.5) \\
\addlinespace

Canidae & Dog (2) &37.1 &37.1 ($\uparrow$0) &56.6 &58.3 ($\uparrow$1.7) &70.0 &71.2 ($\uparrow$1.2) \\
& Fox (26) &34.8 &35.7 ($\uparrow$0.9) &50.3 &51.9 ($\uparrow$1.6) &61.6 &62.2 ($\uparrow$0.6) \\
& Wolf (29) &40.3 &39.5 ($\downarrow$0.8) &57.1 &59.8 ($\uparrow$2.7) &68.7 &69.5 ($\uparrow$0.8) \\
\addlinespace

Felidae & Cat (3) &43.6 &45.0 ($\uparrow$1.4) &62.1 &62.9 ($\uparrow$0.8) &71.8 &73.5 ($\uparrow$1.7) \\
& Tiger (19) &41.0 &43.7 ($\uparrow$2.7) &59.7 &61.2 ($\uparrow$0.5) &68.4 &70.6 ($\uparrow$2.2) \\
& Lion (20) &40.3 &44.2 ($\uparrow$3.9) &59.9 &59.3 ($\downarrow$0.6) &68.6 &69.5 ($\uparrow$0.9) \\
& Cheetah (22) &36.6 &39.2 ($\uparrow$2.6) &56.6 &56.1 ($\downarrow$0.5) &66.1 &67.4 ($\uparrow$1.3) \\
\addlinespace

Equidae & Horse (4) &38.8 &38.9 ($\uparrow$0.1) &63.6 &65.6 ($\uparrow$2.0) &72.8 &74.6 ($\uparrow$1.8) \\
& Zebra (13) &55.9 &58.0 ($\uparrow$2.1) &75.1 &76.0 ($\uparrow$0.9) &81.4 &82.4 ($\uparrow$1.0) \\
\addlinespace

Leporidae & Rabbit (5) &39.3 &41.1 ($\uparrow$1.8) &53.0 &54.7 ($\uparrow$1.7) &63.1 &63.9 ($\uparrow$0.8) \\
\addlinespace

Suidae & Pig (6) &46.9 &49.1 ($\uparrow$2.2) &64.4 &65.5 ($\uparrow$1.1) &71.1 &72.4 ($\uparrow$1.3) \\
\addlinespace

Hominidae & Chimpanzee (7) &33.7 &33.7 ($\uparrow$0) &47.9 &50.6 ($\uparrow$2.7) &58.0 &61.1 ($\uparrow$3.1) \\
& Orangutan (9) &34.2 &35.2 ($\uparrow$1.0) &51.0 &52.9 ($\uparrow$1.9) &59.2 &63.4 ($\uparrow$4.2) \\
& Gorilla (10) &27.7 &27.2 ($\downarrow$0.5) &48.5 &50.7 ($\uparrow$2.2) &56.5 &60.3 ($\uparrow$3.8) \\
\addlinespace

Cercopithecidae & Monkey (8) &39.1 &40.0 ($\uparrow$0.9) &56.9 &58.6 ($\uparrow$1.7) &65.5 &67.5 ($\uparrow$2.0) \\
& Spider Monkey (11) &41.5 &41.6 ($\uparrow$0.1) &53.5 &55.2 ($\uparrow$1.7) &61.9 &64.4 ($\uparrow$2.5) \\
& Night Monkey (12) &39.7 &39.5 ($\downarrow$0.2) &54.0 &54.4 ($\uparrow$0.4) &61.1 &64.3 ($\uparrow$3.2) \\
\addlinespace

Elephantidae & Elephant (14) &51.2 &55.6 ($\uparrow$4.4) &75.1 &74.0 ($\downarrow$1.1) &75.3 &77.3 ($\uparrow$2.0) \\
\addlinespace

Hippopotamidae & Hippo (15) &53.0 &55.3 ($\uparrow$2.3) &72.4 &74.6 ($\uparrow$2.2) &77.2 &78.7 ($\uparrow$1.5) \\
\addlinespace

Procyonidae & Raccoon (16) &42.0 &43.5 ($\uparrow$1.5) &58.0 &58.8 ($\uparrow$0.8) &67.0 &69.7 ($\uparrow$2.7) \\
\addlinespace

Rhinocerotidae & Rhino (17) &49.8 &52.1 ($\uparrow$2.3) &65.9 &67.9 ($\uparrow$2.0) &75.0 &75.6 ($\uparrow$0.6) \\
\addlinespace

Giraffidae & Giraffe (18) &53.4 &56.3 ($\uparrow$2.9) &76.7 &77.8 ($\uparrow$1.1) &80.7 &83.1 ($\uparrow$2.4) \\
\addlinespace

Ursidae & Panda (21) &43.0 &41.7 ($\downarrow$1.3) &58.2 &60.1 ($\uparrow$1.9) &65.8 &71.8 ($\uparrow$6.0) \\
& Black Bear (23) &37.8 &38.8 ($\uparrow$1.0) &54.7 &56.5 ($\uparrow$1.8) &60.4 &62.5 ($\uparrow$2.1) \\
& Polar Bear (24) &44.9 &48.8 ($\uparrow$3.9) &63.0 &64.9 ($\uparrow$1.9) &71.3 &73.4 ($\uparrow$2.1) \\
\addlinespace

Bovidae & Antelope (25) &50.1 &53.5 ($\uparrow$3.4) &72.7 &73.2 ($\uparrow$0.5) &77.5 &77.9 ($\uparrow$0.4) \\
& Buffalo (27) &56.5 &60.7 ($\uparrow$4.2) &68.3 &69.1 ($\uparrow$0.8) &71.5 &73.9 ($\uparrow$2.4) \\
& Cow (28) &52.5 &54.6 ($\uparrow$2.1) &73.5 &73.8 ($\uparrow$0.3) &79.3 &79.8 ($\uparrow$0.5) \\
& Sheep (30) &39.1 &39.6 ($\uparrow$0.5) &63.5 &64.3 ($\uparrow$0.8) &70.9 &72.2 ($\uparrow$1.3) \\

\addlinespace
\midrule
\multicolumn{2}{l}{Std. across species}
& 7.19
& 8.21
& 8.34
& 7.99
& 6.81
& 6.30 \\

\bottomrule
\end{tabular}
}

\caption{\centering Unsupervised results on APT-v2 across 15 families. Values report mean PCK@0.05$_{\text{img}}$ per family. $\uparrow$ indicates performance change relative to the previous column. The row Std. across species reports the standard deviation computed over all 30 species in APT-v2.}
\label{taba3}
\end{table}

\newpage
Table \ref{taba4} reports tracking performance on user-defined intermediate points, demonstrating the flexibility of our framework beyond predefined anatomical landmarks. To enable quantitative evaluation, the ground-truth location of each point is defined as the midpoint of an annotated joint pair and is computed directly from the corresponding endpoint annotations in every frame. Specifically, We evaluate the midpoint between joints 4 and 5 (a spine midpoint) and the midpoint between joints 10 and 11 (the forelimb), representing relatively rigid and articulated regions, respectively.

\begin{table}[h]
\centering
\resizebox{0.75\textwidth}{!}{
\begin{tabular}{lcc}
\toprule
\multirow{2}{*}{\textbf{Intermediate Point}} &
\multicolumn{2}{c}{\textbf{APTv2}} \\
\cmidrule(lr){2-3}
& \textbf{PCK@0.1$_{\text{img}}$} & \textbf{PCK@0.05$_{\text{img}}$} \\
\hline
Midpoint of Joint Pair (4,5) & 92.9 & 77.7 \\
Midpoint of Joint Pair (10,11) & 81.8  & 62.9\\
\bottomrule
\end{tabular}
}
\caption{Tracking performance on user-defined intermediate points. Ground-truth locations are computed as the midpoints of annotated joint pairs in each frame, enabling quantitative evaluation of arbitrary point tracking.}
\label{taba4}
\end{table}

\newpage
\section{Qualitative Analysis of Drift Correction}

\begin{figure}[!h]
	\centering
	\includegraphics[width=\textwidth]{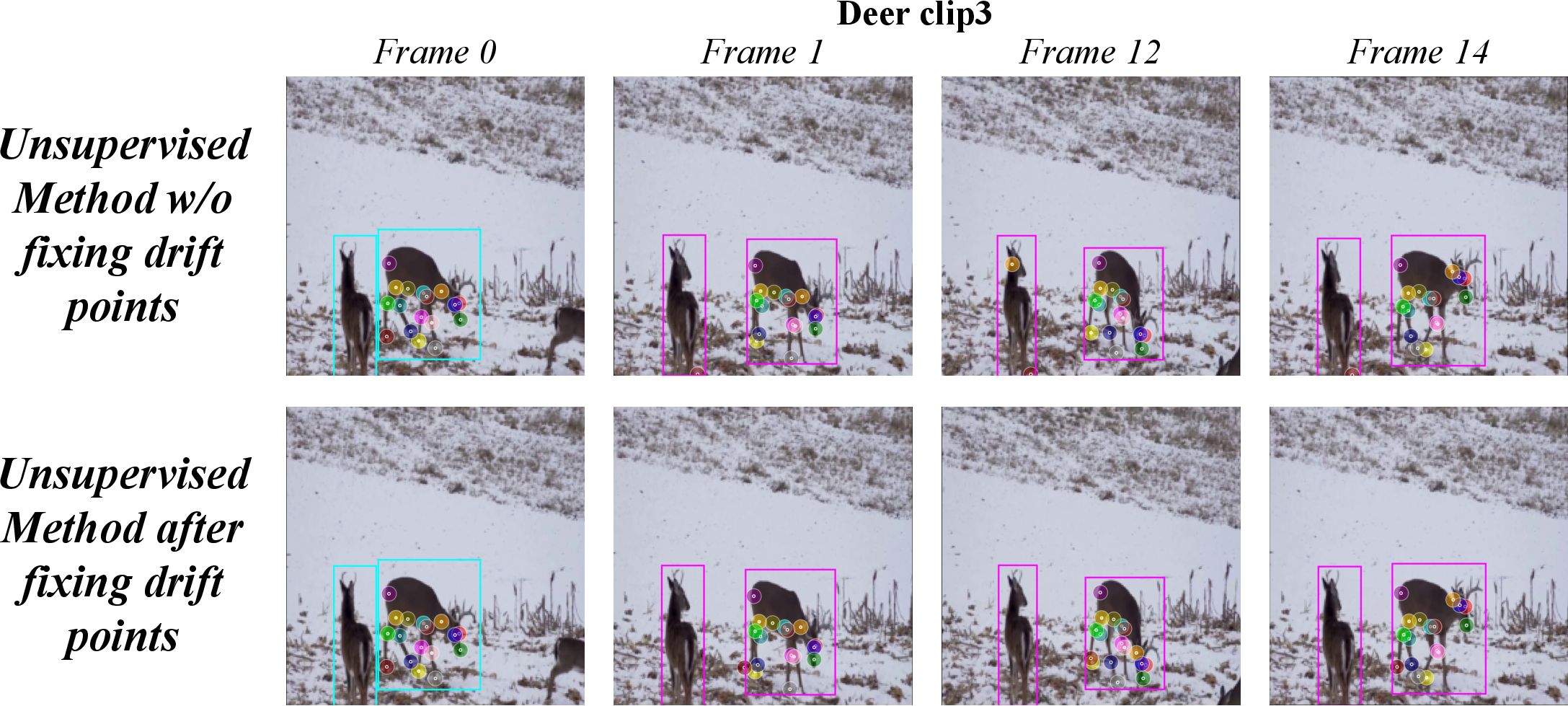}

\caption{
Qualitative results of drift correction in the unsupervised route. 
Top: without drift correction, keypoints may drift across instances in multi-animal scenes. 
Bottom: after applying the proposed correction, cross-instance drift is effectively mitigated and keypoints are reassigned to the correct target animal. 
The bounding boxes are estimated from predicted instance extents rather than ground-truth annotations, consistent with our promptable tracking setting.
}

	\label{figa1}
\end{figure}

Figure~\ref{figa1} provides a qualitative analysis of the proposed drift correction mechanism. 
Without explicit constraints, keypoints in the unsupervised route may drift across instances when multiple animals appear in the same scene, leading to incorrect associations. 
By introducing a bounding-box-based spatial constraint, such drift is effectively mitigated, ensuring that correspondences remain within the correct object region. 
As shown in the figure, keypoints that previously switch to the instance nearby are successfully reassigned back to the target instance after correction. Importantly, the use of dataset-provided bounding boxes is optional. The unsupervised route already achieves promising performance without bounding-box constraints, while incorporating them provides a further improvement by reducing cross-instance drift.

\newpage
\section{Analysis of Temporal Robustness across Frames}

\begin{figure}[!h]
\centering

\begin{subfigure}[t]{0.48\textwidth}
    \centering
    \includegraphics[width=\linewidth]{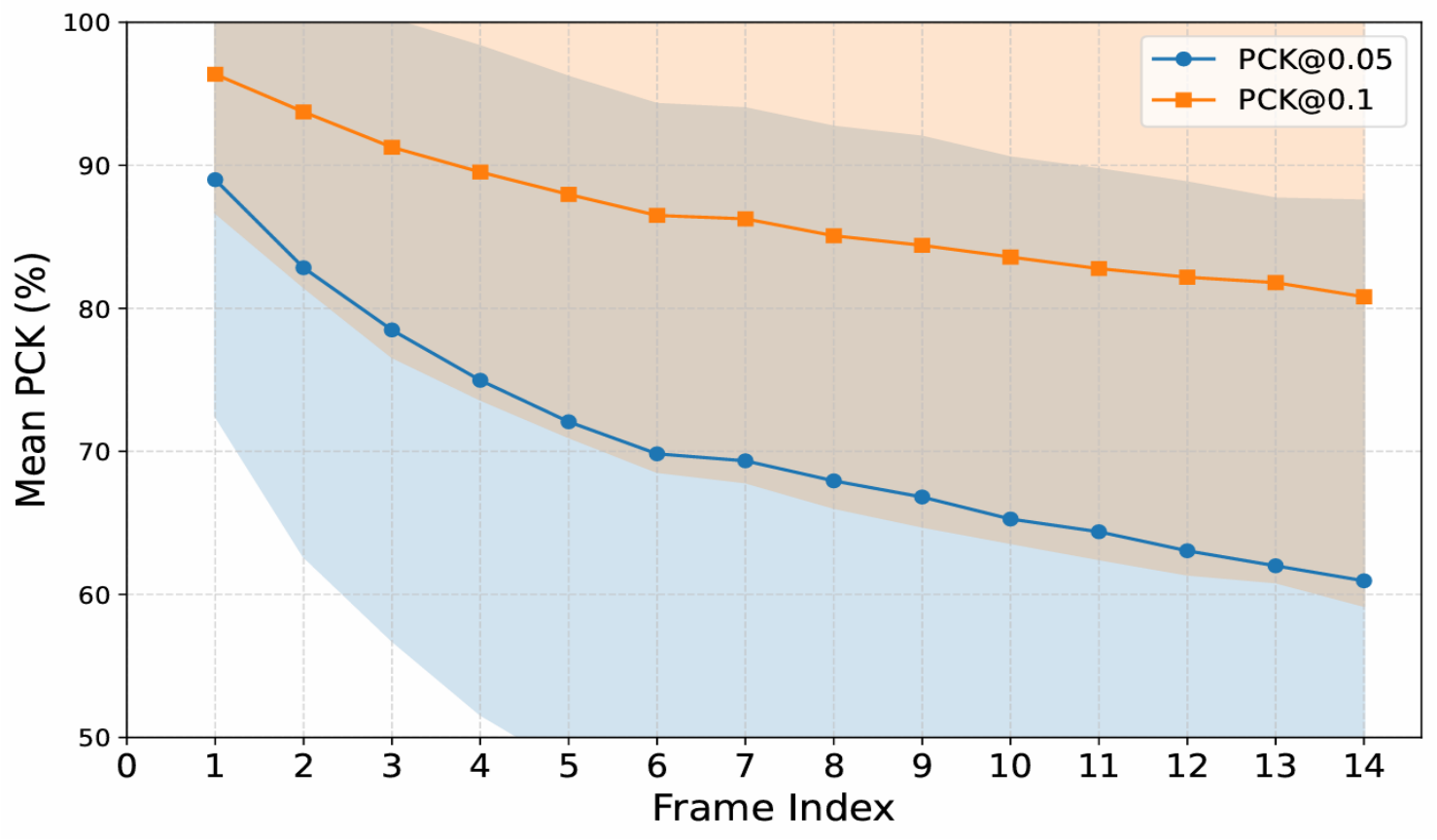}
    \caption{}
    \label{fig:supp1a}
\end{subfigure}
\hfill
\begin{subfigure}[t]{0.48\textwidth}
    \centering
    \includegraphics[width=\linewidth]{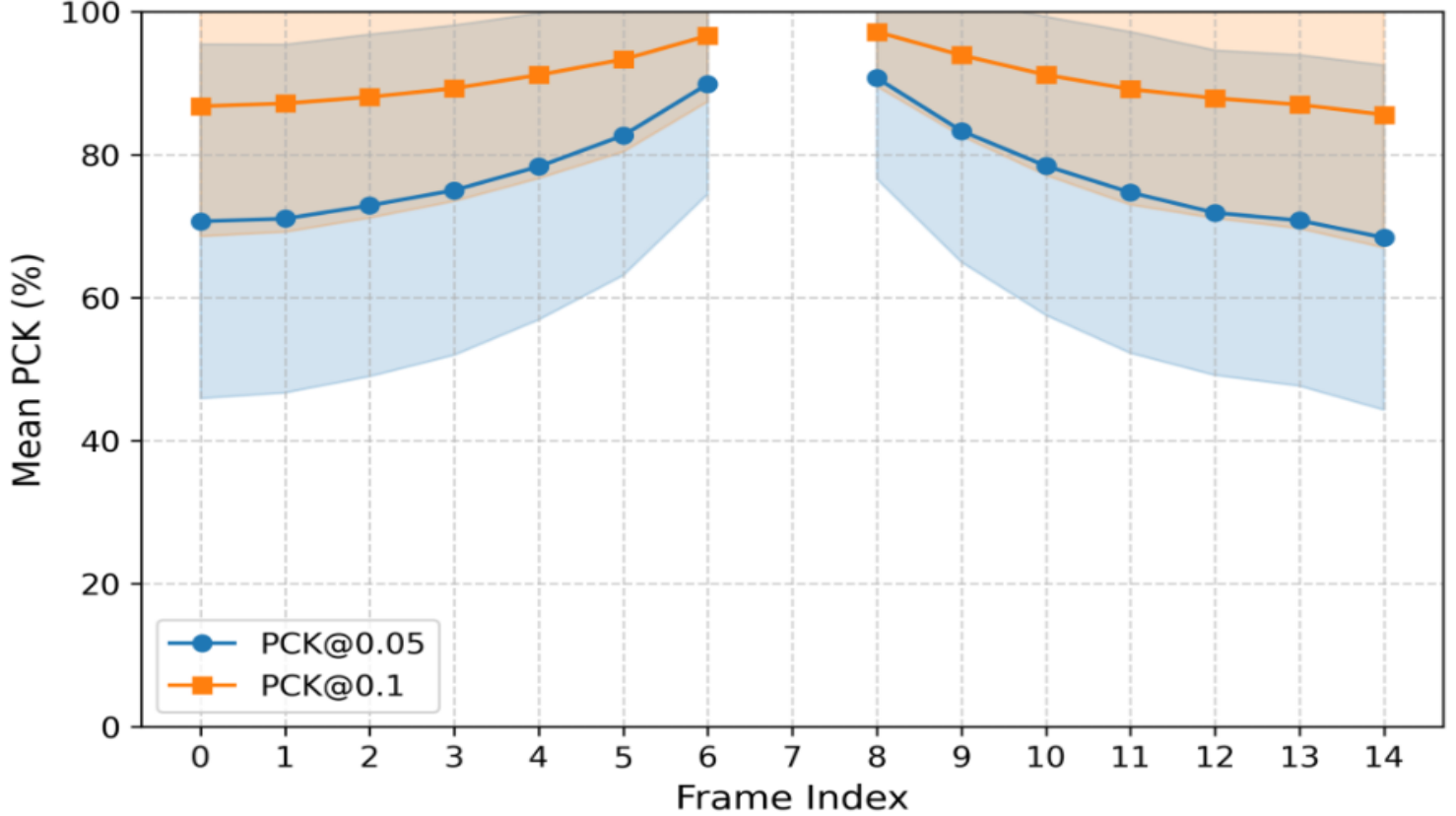}
    \caption{}
    \label{fig:supp1b}
\end{subfigure}

\caption{
Mean PCK@0.1$_{\text{img}}$ and PCK@0.05$_{\text{img}}$ across frame indices on APT-v2 under different reference-frame selections. Error bars indicate the standard deviation across videos. (a) The reference frame is fixed as the first frame ($r=0$). (b) The reference frame is fixed as the middle frame ($r=7$).
}
\label{figa2}
\end{figure}
Figure \ref{figa2} provides a complementary analysis of the influence of reference-frame selection. When the first frame is used as the reference (Figure A1(a)), tracking performance gradually decreases as the temporal distance increases. In contrast, selecting the middle frame as the reference (Figure A1(b)) results in a more symmetric performance profile, with substantially less pronounced degradation in both temporal directions. We hypothesize that this is because the appearance and pose differences between the middle frame and its neighbouring frames are generally smaller than those between the first frame and the rest of the sequence. Consequently, using a middle reference frame can improve the average tracking performance when offline processing is feasible. This observation suggests that reference-frame selection is an important factor for correspondence-based tracking methods and may provide a simple yet effective strategy for improving long-range tracking performance.

\newpage
\section{Qualitative Analysis of Leave-one-out}

\begin{figure}[!h]
	\centering
	\includegraphics[width=\textwidth]{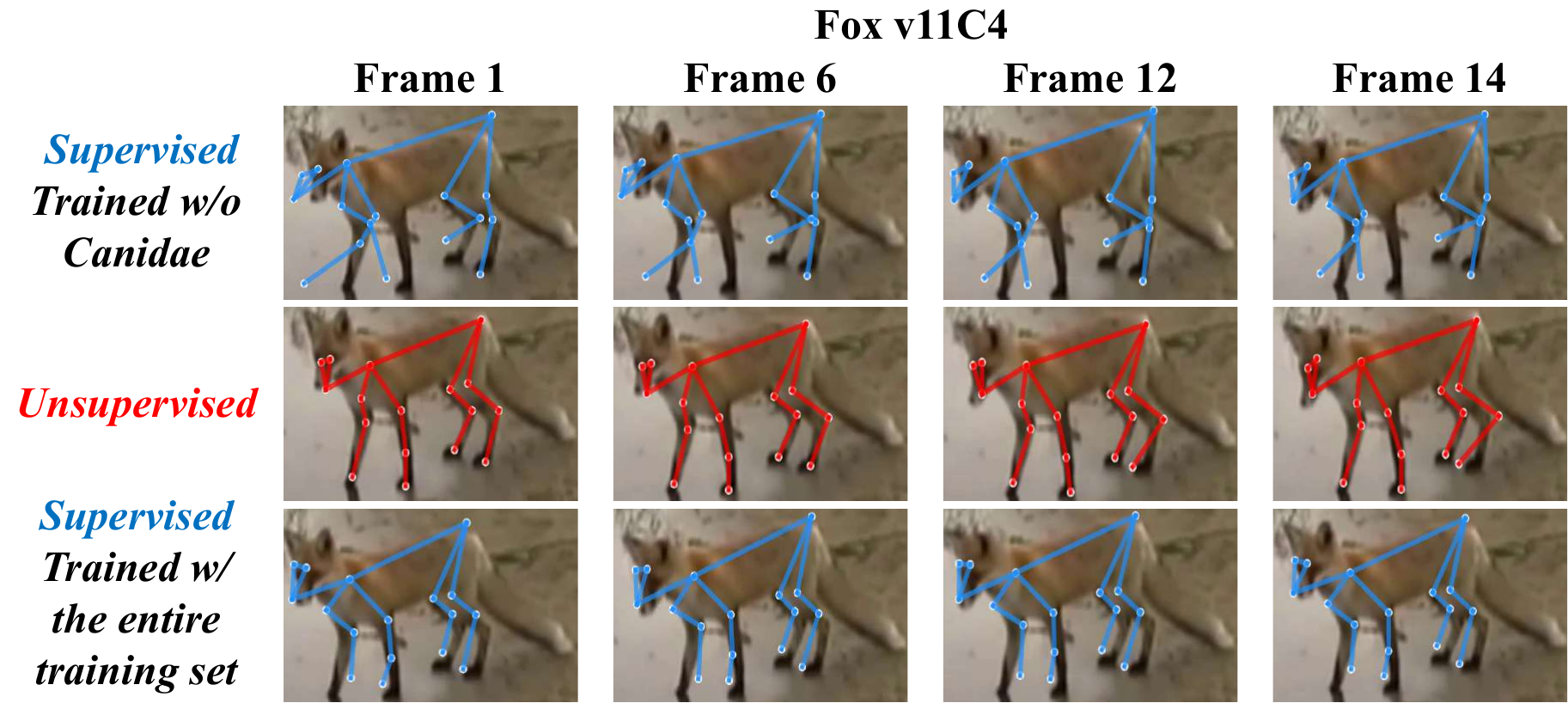}
\caption{
Qualitative results of leave-one-out analysis on fox sequences. 
The results are shown on the same input frames for direct comparison.
}
	\label{figa3}
\end{figure}
As shown in Figure \ref{figa3}, we further analyse the leave-one-out setting on fox sequences under three different training configurations. The model trained without Canidae exhibits noticeably degraded performance, with keypoints often drifting toward semantically similar but incorrect regions, suggesting a bias introduced by the absence of closely related training categories. Interestingly, this setting performs below the unsupervised route, indicating that supervised optimization may overfit to the remaining training distribution when key category priors are missing. In contrast, training with the full dataset yields significantly more stable and accurate predictions. These observations suggest that supervised learning can suffer from category imbalance and negative transfer in cross-family generalization, while unsupervised features provide more consistent baseline correspondence in unseen categories.

\end{document}